\documentclass[10pt,twocolumn,letterpaper]{article}
\usepackage{authblk}
\usepackage{subfigure}
\usepackage{cvpr}                

\usepackage[dvipsnames]{xcolor}

\usepackage{graphicx}

\usepackage{epsfig}
\usepackage{bbding} 
\usepackage{pifont} 
\usepackage{wrapfig} 
\usepackage{microtype}
\usepackage{amsfonts}

\usepackage{amsmath,amssymb} 
\usepackage{mathrsfs} 
\usepackage{algorithmic,algorithm}
\usepackage{url} 
\usepackage{multirow}  
\usepackage{pifont} 
\usepackage{microtype} 
\usepackage{array}
\usepackage{soul}
\usepackage{booktabs}
\usepackage{caption,subcaption}
\usepackage{colortbl}

\newtheorem{theorem}{Theorem}

\newcommand{\cmark}{\ding{51}}%
\newcommand{\xmark}{\ding{55}}%

\def\modelshortname{DIFO}

\definecolor{cmblu}{RGB}{51,102,240}
\definecolor{cmred}{RGB}{241,22,22}

\definecolor{cvprblue}{rgb}{0.21,0.49,0.74}
\usepackage[pagebackref,breaklinks,colorlinks,citecolor=cvprblue]{hyperref}

\def\paperID{10888} 
\def\confName{CVPR}
\def\confYear{2024}

\title{Source-Free Domain Adaptation with Frozen Multimodal Foundation Model}
\author[1,2,3]{Song Tang}
\author[1]{Wenxin Su}
\author[4]{Mao Ye\thanks{Corresponding author}}
\author[*5]{Xiatian Zhu}
\vspace{-10pt}
\affil[1]{University of Shanghai for Science and Technology  \textsuperscript{2}Universität Hamburg \textsuperscript{3}ComOriginMat Inc.}
\affil[4]{University of Electronic Science and Technology of China \textsuperscript{5}University of Surrey}
\affil[ ]{
{\tt\small tangs@usst.edu.cn, 
\{suwenxin43, cvlab.uestc\}@gmail.com
, xiatian.zhu@surrey.ac.uk}}

\begin{document}
\maketitle
\begin{abstract}

Source-Free Domain Adaptation~(SFDA) aims to adapt a source model for a target domain, with only access to unlabeled target training data and the source model pre-trained on a supervised source domain. 
Relying on pseudo labeling and/or auxiliary supervision, conventional methods are inevitably error-prone.
To mitigate this limitation, in this work we for the first time explore the potentials of off-the-shelf vision-language (ViL) multimodal models (e.g., CLIP) with rich whilst heterogeneous knowledge.
We find that directly applying the ViL model
to the target domain in a zero-shot fashion is unsatisfactory, as it is not specialized for this particular task but largely generic. 
To make it task specific, we propose a novel \textit{\textbf{D}istilling mult\textbf{I}modal \textbf{F}oundation m\textbf{O}del}~(\textbf{\modelshortname}) approach. 
Specifically, \modelshortname{} alternates between two steps during adaptation:
(i) Customizing the ViL model by maximizing the mutual information with the target model in a prompt learning manner,
(ii) Distilling the knowledge of this customized ViL model to the target model.
For more fine-grained and reliable distillation, we further introduce two effective regularization terms, namely most-likely category encouragement and predictive consistency.
Extensive experiments show that {\modelshortname} significantly outperforms the state-of-the-art alternatives. 
Code is \href{https://github.com/tntek/source-free-domain-adaptation}{here}.
\end{abstract}

\section{Introduction}
\label{sec:intro}

Unsupervised Domain Adaptation (UDA) relies on both well-annotated source data and unannotated target data. However, due to heightened safety and privacy concerns, accessing source data freely has become difficult \cite{2019Distant, lao2021hypothesis}. 
In response, Source-Free Domain Adaptation (SFDA) has gained attention as a more practical solution, aiming to transfer a pre-trained source model to the target domain using only unlabeled target data. 


\begin{figure}[t] 
    \setlength{\belowcaptionskip}{-5pt}
    \setlength{\abovecaptionskip}{-2pt}
    \begin{center}
     \includegraphics[width=0.9\linewidth]{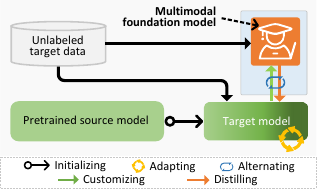}
    \end{center}
    \caption{
    We expand beyond traditional SFDA methods that rely solely on a pretrained source model and unlabeled target data. Instead, we innovate by exploring off-the-shelf multimodal foundation models, such as CLIP, in an unsupervised manner (marked by the box with blue background).
    }
    \label{fig:idea-comp}
\end{figure}

Due to the absence of source samples, traditional distribution matching approaches are no longer viable~\cite{ganin2015unsupervised,kang2019contrastive}. The predominant alternative is self-supervised learning, which generates or mines auxiliary information to facilitate unsupervised adaptation. Two main approaches exist: constructing a pseudo source domain to leverage established UDA methods such as adversarial learning~\cite{xia2021adaptive,kurmi2021domain} or domain shift minimization based on distribution measurement~\cite{ding2022source,tian2021vdm,kundu2022balancing} and mining extra supervision from the source model~\cite{lao2021hypothesis,wang2022exploring,huang2021model} or target data~\cite{yang2022attracting,tang2022sclm,yang2021nrc}.
In the presence of domain distribution shift, applying the source model to the target domain introduces inevitable errors in pseudo-labeling or auxiliary supervision, thereby limiting adaptation performance.

To address identified limitations, we pioneer the exploration of off-the-shelf multimodal foundation models, such as the vision-language (ViL) model CLIP~[24], transcending the constraints of both the source model and target data knowledge. 
However, direct application of the ViL model proves unsatisfactory, lacking specialization for specific tasks. 
To overcome this, we propose a novel task-specific distillation approach named \textit{\textbf{D}istilling mult\textbf{I}modal \textbf{F}oundation m\textbf{O}del~(\textbf{\modelshortname})}.
Initially, we customize the ViL model through {\em unsupervised} prompt learning for imposing task-specific information.
Subsequently, we distil the knowledge from this customized ViL model to the target model, with joint supervision through two designed regularization terms: (1) most-likely category encouragement for coarse-grained distillation and (2) predictive consistency for fine-grained distillation.

Our \textbf{contributions} are summarized as follows. 
\textbf{(1)}~Pioneering the use of generic but heterogeneous knowledge sources (e.g., the off-the-shelf ViL model) for the SFDA problem, transcending the limited knowledge boundary of a pretrained source model and unlabeled target data.
\textbf{(2)}~Development of the novel {\modelshortname} approach to effectively distill useful task-specific knowledge from the general-purpose ViL model.
\textbf{(3)}~Extensive evaluation on standard benchmarks, demonstrating the significant superiority of our {\modelshortname} over previous state-of-the-art alternatives under conventional closed-set settings, as well as more challenging partial-set and open-set settings.

\section{Related Work}\label{sec:rework}

\begin{figure*}[t]
    \setlength{\belowcaptionskip}{-5pt}
    \setlength{\abovecaptionskip}{-1pt}
    \begin{center}
        \includegraphics[width=0.9\linewidth,angle=0]{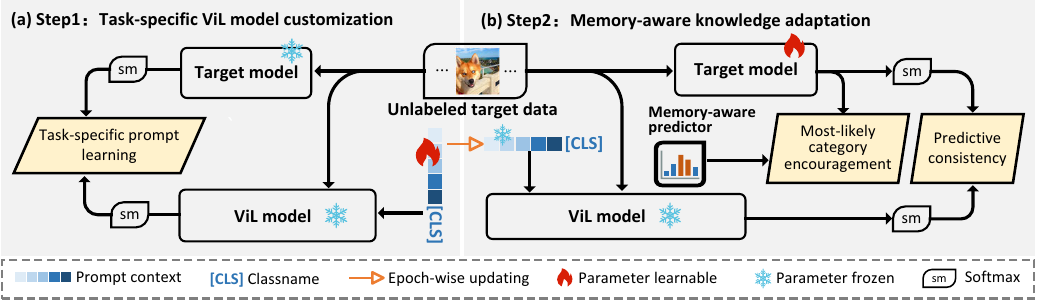}
    \end{center}
    \caption{
    Overview of our {\modelshortname}: The process involves two alternating steps. First, we perform (a) {\em task-specific customization} of a ViL model through task-specific prompt learning ($L_{\rm{TSC}}$). This is achieved under soft predictive guidance using mutual information maximization. Second, we undertake (b) {\em memory-aware knowledge adaptation}, incorporating two regularizations: most-likely category encouragement ($L_{\rm{MCE}}$) predicted by our dynamic memory-aware predictor, along with the tupical predictive consistency ($L_{\rm{PC}}$). These regularizations are designed to facilitate a coarse-to-fine adaptation.
    }
    \label{fig:fw}
\end{figure*}

\textbf{Source-free domain adaptation.}
Existing SFDA approaches fall into three distinct categories. The first explicitly aligns the pseudo source domain with the target domain, treating SFDA as a specialized case of unsupervised domain adaptation. This alignment is achieved by constructing the pseudo source domain through a generative model~\cite{2020MA, tian2022vdm} or by splitting the target domain based on prior source hypotheses~\cite{du2021ps}.

The second group extracts cross-domain factors from the source domain and transfers them in successive model adaptation for aligning feature distributions across the two domains. For example, \cite{tang2019adaptive} establishes a mapping relationship from a sample and its exemplar Support Vector Machine (SVM) (an individual classifier) on the source domain to ensure individual classification on the target domain. Some approaches leverage pre-trained source models to generate auxiliary factors, such as multi-hypothesis~\cite{lao2021hypothesis}, prototypes~\cite{tanwisuth2021pct}, source distribution estimation~\cite{ding2022source}, or hard samples~\cite{li2021divergence} to aid in feature alignment.

The third group incorporates auxiliary information refined from the unlabeled target domain. In addition to widely used pseudo-labels~\cite{liang2020we, chen2022self}, geometry information, such as intrinsic neighborhood structure~\cite{tang2021nearest} and target data manifold~\cite{tang2022sclm}, has also been exploited.

Despite continual advancements, these methods are limited by the knowledge derived solely from the pretrained source model and unlabeled target data. We break this limitation by tapping into the rich knowledge encoded in off-the-shelf multimodal foundation models.

\vspace{2pt}
\noindent\textbf{Large multimodal model.}
Multimodal vision-language (ViL) models, such as CLIP~\cite{radford2021learning} and ALIGN~\cite{jia2021scaling}, have shown promise across various mono-modal and multimodal tasks by capturing modality-invariant features. Approaches in this domain can be broadly categorized into two lines.

The first line focuses on enhancing ViL model performance. For instance, in~\cite{zhou2022learning,ge2022domain}, prompt learning optimizes the text encoder through the use of tailored, learnable prompts designed for specific scenarios. Other efforts aim to improve data efficiency by repurposing noisy data~\cite{andonian2022robust}.

The second line utilizes ViL models as external knowledge to enhance downstream tasks, as demonstrated in this paper. Previous work in knowledge transfer primarily falls into two frameworks. For the first scheme, where the ViL model is directly applied to the target task in a zero-shot fashion~\cite{liang2023open}, domain generality is leveraged without task-specific refinement. The second scheme does not focus on source model adaptation. Instead, it fine-tunes the ViL model to the target domain through prompt or adaptor learning with an amount of manal labels~\cite{Cho_2023_ICCV}.

A relevant method to our {\modelshortname} is the UDA method DAPL~\cite{ge2022domain}. Although both adopt CLIP, they differ significantly in problem setting and methodology. DAPL employs CLIP to learn domain-specific prompts, aiming to disentangle domain and category information in CLIP's visual features. In contrast, {\modelshortname} aligns target features to a progressively customized vision-language latent space in a memory-aware fashion. Importantly, DAPL requires labeled source data, making it inapplicable in SFDA.

\section{Methodology}\label{sec:method}
\textbf{Problem statement.} 
In the context of two distinct yet interrelated domains—namely, the labeled source domain and the unlabeled target domain—both characterized by the same set of $C$ categories, the following notation is employed. The source samples and their corresponding labels are represented as $\mathcal{X}_s$ and $\mathcal{Y}_s$ respectively. Similarly, the target samples and their true labels are denoted as $\mathcal{X}_t\!=\!\{{\boldsymbol{x}_{i}\}_{i=1}^{n}}$ and $\mathcal{Y}_t\!=\!\{{y}_{i}\}_{i=1}^{n}$, where $n$ signifies the number of samples.

We aim to learn a target model $\theta_t\!:\!\mathcal{X}_t\! \to \!\mathcal{Y}_t$. This involves utilizing (1) a pre-trained source model $\theta_s\!:\!\mathcal{X}_s \!\to \!\mathcal{Y}_s$, (2) unlabeled target data, and (3) a Visual-Language (ViL) model denoted as ${{\theta}}_{v}$.

\vspace{2pt}
\noindent\textbf{Overview.}
As depicted in Fig.~\ref{fig:fw}, the proposed {\modelshortname} framework alternates between two distinct steps to customize and distill the off-the-shelf ViL knowledge.  

\textit{In the first step}, we engage in prompt learning on the ViL model for the purpose of task-specific customization. This serves to mitigate the guidance error within the ViL model. In particular, we adopt a mutual information-based alignment approach. This approach is characterized by its richness in context and interaction between the target model and the ViL model, as opposed to placing blind trust in either model alone as conventional methods.



\textit{In the second step}, knowledge adaptation takes place within a unique constraint that encourages the identification of the most probable category labels in the logit space, while concurrently maintaining the typical predictive consistency. The most likely category labels are determined by a carefully designed memory-aware predictor, which dynamically integrates knowledge from both the target model and the ViL model in a cumulative fashion.

\subsection{Task-Specific ViL Model Customization} \label{sec:tsc}

We adopt the prompt learning framework for ViL model customization,
with all the parameters of the ViL model frozen throughout.
The only learnable part in customization is the prompts each assigned for a specific class.
To optimize these prompts, we need a useful supervision.
In SFDA, however, it is challenging to customize such a domain-generic ViL model towards to the target domain, at the absence of a well-trained target domain model.
This is because, none of them can reasonably make predictions.
That means there is no clearly good supervision signals available.

To address this challenge, we propose to explore the wisdom of the crowd
by leveraging their predictive interaction as the supervision.
Formally, we denote the predictions by the target model and the ViL model as $\theta_{t}\left({\boldsymbol{x}}_{i}\right)$ and $\theta_{v}\left({\boldsymbol{x}}_{i}\right)$, respectively, given an unlabeled target sample ${\boldsymbol{x}}_{k}$.
We conduct the customization by maximizing the mutual information of their predictions as:
\begin{equation}
    \label{eqn:loss_mim}
    \begin{aligned}
        L_{\rm{TSC}} =- \min_{\boldsymbol{v}}{\mathbb{E}_{{\boldsymbol{x}}_{i} \in {\mathcal{X}_t}}}{{\rm{I}}}\left(\theta_{t}\left({\boldsymbol{x}}_{i}\right), {\theta_{v}} \left(\boldsymbol{x}_{i}, \boldsymbol{v} \right)\right)
    \end{aligned}
\end{equation}
where $\boldsymbol{v}$ is the prompt context to be learned and the function ${{\rm{I}}}(\cdot,\cdot)$ measures the mutual information~\cite{ji2019invariant}. 

This alignment design differs significantly from the conventional adoption of the Kullback–Leibler (KL) divergence. 
First of all, the mutual information is a lower optimization bound than KL divergence, facilitating deeper alignment (see Theorem~\ref{thm-one} with the proof provided in \texttt{Supplementary}). 
\begin{theorem}
\textit{Given two random variables $X$, $Y$. Their mutual information ${\rm{I}}\left( X, Y \right)$ and KL divergence $D_{\rm{KL}}\left( X||Y \right)$ satisfy the unequal relationship as follows.} 
\begin{equation}
    \label{eqn:dsib}
    -{\rm{I}}\left( X, Y \right) \leq D_{\rm{KL}}\left( X, Y \right). 
\end{equation} 
\label{thm-one} 
\end{theorem}
\vspace{-15pt}

Crucially, the KL divergence exhibits an inherent bias towards a specific prediction, making it less suitable for our context where none of the predictions holds a significant advantage. On the contrary, mutual information considers the joint distribution or correlation between the two predictions. This distinction arises from their respective definitions:
$-{\rm{I}}\left( X, Y \right)=-H\left( X\right)+H\left( X|Y \right)$ and $D_{\rm{KL}}\left( X, Y \right)=-H\left( X\right)+H\left( X:Y \right)$, where
\begin{equation}
    \label{eqn:discuss-bias}
    \begin{split}
        H\left( X~|~Y \right) &=-\sum p(\boldsymbol{x},\boldsymbol{y})\log p(\boldsymbol{x} | \boldsymbol{y})\\
        H\left( X:Y \right)   &=-\sum p(\boldsymbol{x})\log p(\boldsymbol{y}).
    \end{split}
\end{equation}

The conditional entropy component $H\left( X|Y \right)$ of mutual information explicitly captures the joint distributions, a feature absent in KL divergence. Empirically, we also confirm the significance of incorporating this joint distribution-based interaction between the two predictions during the customization of the ViL model (see ablation study in \texttt{Tab.\ref{tab:ab_loss}} and task-specific knowledge adaptation analysis in \texttt{Section}~\ref{sec:tskaa}).

\subsection{Memory-Aware Knowledge Adaptation}

As previously mentioned, even with customization for the target domain, the ViL model may not be fully adapted due to no robust target model available in prior. This limitation hinders effective knowledge adaptation at this stage. To address this issue, we propose the incorporation of a specialized memory-aware predictor to provide additional learning guidance -- most-likely category encouragement, complementing the conventional predictive consistency constraint.

\begin{figure}[t]
    \begin{center}
	\includegraphics[width=0.95\linewidth]{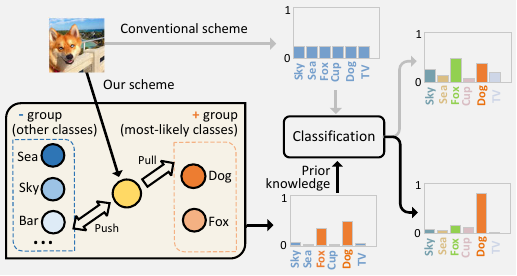}
    \end{center}
    \setlength{\abovecaptionskip}{0cm}
    \setlength{\belowcaptionskip}{-0.3cm}
    \caption{
    Illustration of most-likely category encouragement. 
    In contrast to the conventional approach that assigns equal importance to all categories (depicted by the gray line), our approach (represented by the black line) introduces additional supervision by incorporating extra knowledge about the two most likely categories.
    } 
    \label{fig:class-atten}
\end{figure}

\vspace{3pt}
\noindent\textbf{Most-likely category encouragement.}
The rationale behind incorporating this learning constraint is to harness the collective knowledge of both the target model and the ViL model in order to enhance the discernment of probable category labels for each sample. Given the sluggish nature of this search process, it has been devised to function as a form of learning regularization. 
An illustration of this regularization process is presented in Fig.~\ref{fig:class-atten}.
Specifically, it is realized through two distinct steps as detailed below.

{\bf \textit{(I) Memory-aware predictor.}}
We initiate the process by generating pseudo-labels that represent the most likely category distribution, utilizing historical information stored in a prediction bank. The prediction bank archives two types of historical data for all samples in the target domain: (1) predictions from the target model denoted by $\{\boldsymbol{p}_i\}_{i=1}^{n}$ and (2) predictions from the ViL model denoted by $\{\boldsymbol{p}'_i\}_{i=1}^{n}$.

Throughout the adaptation process, the predictions from the target model are updated iteratively. At the end of each training iteration, the newly predicted labels for the training batch from the target model replace their counterparts in the prediction bank. In contrast, predictions from the ViL model are updated collectively in an epoch-wise manner, triggering updates every $M$ iterations. This mixed-update strategy is designed to strike a balance between maintaining the stability of the customized ViL model's guidance and capturing the task-specific dynamics inherent in the adaptation process.

Based on the provided prediction bank, the creation of a pseudo-label for the most probable category involves a historical prediction fusion process as:
\begin{equation}
    \label{eqn:onehot-fusion}
    \begin{split}
        {\boldsymbol{\bar{p}}}_{i} = \omega~{\boldsymbol{p}}_{i} + (1-\omega)~\boldsymbol{p}'_{i}.
    \end{split}
\end{equation}
Here, the weight $\omega$, drawn from an Exponential distribution with parameter $\lambda$, is a crucial factor. This fusion introduces dynamic bias rectification (represented by ${\boldsymbol{p}}_{i}$) based on the guidance from the customized ViL model ($\boldsymbol{p}'_{i}$). The role of ${\boldsymbol{p}}_{i}$ is to provide adjustments, leading us to adopt an asymmetric random weighting approach represented by $\omega$.

{\bf \textit{(II) Category attention calibration.}}
Subsequently, we formulate a regularization technique employing pseudo-labels acquired through category attention calibration. Specifically, we begin by identifying the {\it top-N} most probable categories using ${\boldsymbol{\bar{p}}}_{i}$. The indices of these identified categories are denoted by $\mathcal{M}_i=\{m_k\}_{k=1}^{N}$. With $\mathcal{M}_i$, the target model's logit of a target domain sample $x_i$, denoted as $\boldsymbol{l}_i$, is segregated into positive and negative category groups.
We define this regularization as:
\begin{equation}\label{eqn:loss_mce} 
    \begin{split} 
    {L}_{\rm{MCE}} &= \min_{\theta{t}}\mathbb{E}_{{\boldsymbol{x}}_{i} \in {\mathcal{X}_t}}  
    \log \frac{\exp\left( a_i / \tau \right)} {\sum_{\substack{j \neq \mathcal{M}_i}}
    \exp{\left( b_i \cdot \boldsymbol{l}_{i,j}/\tau \right)}}\\
    a_i &= \prod\limits_{k=1}^{N} {\boldsymbol{l}_{i,m_k}}, ~~~
    b_i = \sum\limits_{k=1}^{N} {\boldsymbol{l}_{i,m_k}}
    \end{split}
\end{equation}
where $\boldsymbol{l}_{i, a}$ denotes the $a$-th element of $\boldsymbol{l}_{i}$ and $\tau$ is the temperature parameter. 

In Eq.~\eqref{eqn:loss_mce}, we note that the product operation with $a_i$ in the numerator amplifies penalties for the probability decrease on the most likely categories compared to the sum form. Similarly, the sum with $b_i$ in the denominator serves as an increasing weighting parameter to enhance suppression of values at other locations. Moreover, $a_i$ is more sensitive to changes than $b_i$ due to $\frac{{\partial a_i}}{{\partial m_k}} \propto {O}(n^{N-1})$ and $\frac{\partial b_i}{\partial m_k} \propto {O}(1)$. By combining the use of $a_i$ and $b_i$, we globally impose a calibration effect on the elements corresponding to the most likely categories within the logit $\boldsymbol{l}_i$.
Essentially, attention is introduced to these potential categories, as illustrated in the box with a yellow background in Fig.~\ref{fig:class-atten}.

\begin{algorithm}[t]
    \caption{Training of {\modelshortname}}
    \label{alg:algorithm}
    \raggedright
    \textbf{Input}: Pre-trained source model $\theta_s$, target model $\theta_t$, ViL model $\theta_{v}$, unlabelled target domain $\mathcal{X}_t$, learnable prompt context $\boldsymbol{v}$, \#epoch $T$, \#iteration per epoch $M$.\\
    \textbf{Output}: The adapted target model $\theta_t$.\\
    \textbf{Procedure}:
    \begin{algorithmic}[1] 
    \STATE \textbf{Initialisation}: Set $\theta_t = \theta_s$ and $\boldsymbol{v}$=\textit{'a photo of a [CLS].'}
    \FOR{$t$ = 1:$T$}
    \STATE Update ViL predictions in the prediction bank.
    \STATE =========== \textit{Step1} ===========
    \FOR{$m$ = 1:$M$} 
    \STATE Sample a batch from $\mathcal{X}_t$;
    \STATE Forward prompt $\boldsymbol{v}$ and this batch $\mathcal{X}_t^b$ through $\theta_{v}$; 
    \STATE Forward this batch data through $\theta_t$; 
    \STATE Customize $\theta_{v}$ by optimizing $L_{\rm{TSC}}$ (Eq. \eqref{eqn:loss_mim}) and obtain task-specific prompt context $\boldsymbol{v}^*$.
    \ENDFOR
    \STATE =========== \textit{Step2} ===========
    \FOR{$m$ = 1:$M$}
    \STATE Sample a batch from $\mathcal{X}_t$;
    \STATE Forward the $\boldsymbol{v}^*$ and this batch through $\theta_{v}$;
    \STATE Forward this batch data through $\theta_t$; 
    \STATE Discover most-likely category (Eq.~\eqref{eqn:onehot-fusion}); 
    \STATE Update model $\theta_t$ by optimizing ${L_{{\text{MKA}}}}$ (Eq.~\eqref{eqn:loss-ka}).
    \STATE Update target predictions in the prediction bank.
    \ENDFOR
    \STATE Set $\boldsymbol{v} =\boldsymbol{v}^*$.
    \ENDFOR
    \STATE \textbf{return} Adapted model $\theta_t$.
    \end{algorithmic}
\end{algorithm}

\vspace{4pt}
\noindent\textbf{Predictive Consistency.}
For the purpose of knowledge adaptation, we incorporate the conventional predictive consistency loss as:
\begin{small} 
\begin{equation}
    \label{eqn:loss_pc}
    \begin{aligned}
        {L}_{\rm{PC}} =\min_{\theta_{t}} \left[- \mathbb{E}_{{\boldsymbol{x}}_{i} \in {\mathcal{X}_t}}{{\rm{I}}}\left(\theta_{t}\left({\boldsymbol{x}}_{i}\right), {\theta_{v}} \left(\boldsymbol{x}_{i}, \boldsymbol{v}*\right)\right)+\alpha {L}_B \right], 
    \end{aligned}
\end{equation}
\end{small} 
where $\theta_{t}(\boldsymbol{x}_{i})$ represents the target prediction, $\theta_{v}(\boldsymbol{x}_{i},\boldsymbol{v})$ denotes the ViL prediction, and $\boldsymbol{v}$ is the prompt context learned during the initial phase of task-specific customization. The function ${{\rm{I}}}(\cdot,\cdot)$ corresponds to the mutual information function. The parameter $\alpha$ serves as a trade-off parameter, and the category balance term ${L}_{\rm{B}} ={\rm{KL}} (\left. {\bar{\boldsymbol{q}}} \right|| {\boldsymbol{\frac{1}{C}}})$ aligns with previous approaches~\cite{yang2021nrc,tang2023source}, preventing solution collapse by ensuring the empirical label distribution $\bar{\boldsymbol{q}}$ matches the uniform distribution $\boldsymbol{\frac{1}{C}}$.
For the reasons elaborated in \texttt{Section~\ref{sec:tsc}}, we employ mutual information for alignment.

\subsection{Model training}  
To systematically distill and leverage task-specific knowledge from the ViL model, we adopt an epoch-wise training approach for {\modelshortname}. The training process is divided into $T$ epochs, each comprising two stages aligned with the two steps in the {\modelshortname} framework (Fig.~\ref{fig:fw}). During the first stage, training is governed by the objective ${L_{{\text{TSC}}}}$, and in the subsequent second stage, the objective function transitions to
\begin{equation} 
    \label{eqn:loss-ka}
    {L_{{\text{MKA}}}} = {L_{{\text{PC}}}} + \beta {L_{{\text{MCE}}}},
\end{equation}
where $\beta$ is a trade-off parameter.
We summarize the whole training procedure of {\modelshortname} in Algorithm~\ref{alg:algorithm}.

\section{Experiments} \label{sec:rlt}
\begin{table}[t]
    \caption{Closed-set SFDA on \textbf{Office-31} (\%)}
    \label{tab:oc}
    \renewcommand\tabcolsep{2.5pt}
    \renewcommand\arraystretch{0.95}
    \scriptsize
    \centering
    \begin{tabular}{ l l | c c c c c c c}
        \toprule
        Method &Venue  &A$\to$D &A$\to$W &D$\to$A &D$\to$W &W$\to$A &W$\to$D &Avg.\\
        \toprule
        Source        &--       &79.1 &76.6 &59.9 &95.5 &61.4 &98.8 &78.6 \\
        \midrule 
        SHOT~\cite{liang2020we}       &ICML20  &93.7 &91.1 &74.2 &98.2 &74.6 &\textbf{\color{cmred}100.} &88.6 \\
        NRC~\cite{yang2021nrc}        &NIPS21   &96.0 &90.8 &75.3 &99.0 &75.0 &\textbf{\color{cmred}100.} &89.4 \\
        GKD~\cite{tang2021model}      &IROS21  &94.6 &91.6 &75.1 &98.7 &75.1 &\textbf{\color{cmred}100.} &89.2 \\

        HCL~\cite{huang2021model}          &NIPS21  &94.7 &92.5 &75.9 &98.2 &77.7 &\textbf{\color{cmred}100.} &89.8 \\
        AaD~\cite{yang2022attracting}      &NIPS22  &96.4 &92.1 &75.0 &\textbf{\color{cmred}99.1} &76.5 &\textbf{\color{cmred}100.} &89.9 \\
        AdaCon~\cite{chen2022contrastive}  &CVPR22  &87.7 &83.1 &73.7 &91.3 &77.6 &72.8 &81.0 \\
        CoWA~\cite{lee2022confidence}      &ICML22  &94.4 &95.2 &76.2 &98.5 &77.6 &99.8 &90.3 \\
        SCLM~\cite{tang2022sclm}      &NN22   &95.8 &90.0 &75.5 &98.9 &75.5 &99.8 &89.4 \\
        ELR~\cite{yi2023source}       &ICLR23 &93.8 &93.3 &76.2 &98.0 &76.9 &\textbf{\color{cmred}100.} &89.6 \\
        PLUE~\cite{Litrico_2023_CVPR} &CVPR23 &89.2 &88.4 &72.8 &97.1 &69.6 &97.9 &85.8 \\
        TPDS~\cite{tang2023source}    &IJCV23 &97.1 &94.5 &75.7 &98.7 &75.5 &99.8 &90.2 \\
        \rowcolor{gray! 40} {\textbf{\modelshortname}-C-RN} &-- &93.6 &92.1 &78.5 &95.7 &78.8 &97.0 &89.3 \\
        \rowcolor{gray! 40} {\textbf{\modelshortname}-C-B32}  &-- &\textbf{\color{cmred}97.2}  &\textbf{\color{cmred}95.5} &\textbf{\color{cmred}83.0} &97.2 &\textbf{\color{cmred}83.2} &98.8 &\textbf{\color{cmred}92.5} \\
        \bottomrule
    \end{tabular}
\end{table}

\begin{table*}[t]
    \caption{Closed-set SFDA on \textbf{Office-Home} and \textbf{VisDA}~(\%). 
    \textbf{SF} and \textbf{M} means source-free and multimodal, respectively; 
    the full results on \textbf{VisDA} are in \texttt{Supplementary}.}
    \label{tab:oh}
    \renewcommand\tabcolsep{2.3pt}
    \renewcommand\arraystretch{0.9}
    \scriptsize
    \centering
    \begin{tabular}{ l l | c c | c c c c c c c c c c c c c | c}
        \toprule
        \multirow{2}{*}{Method} &\multirow{2}{*}{Venue} 
        &\multirow{2}{*}{\textbf{SF}}
        &\multirow{2}{*}{\textbf{M}}
        &\multicolumn{13}{c}{\textbf{Office-Home}}\vline
        &{\textbf{VisDA}} \\
        & & & &Ar$\to$Cl &Ar$\to$Pr &Ar$\to$Rw
        &Cl$\to$Ar &Cl$\to$Pr &Cl$\to$Rw    
        &Pr$\to$Ar &Pr$\to$Cl &Pr$\to$Rw  
        &Rw$\to$Ar &Rw$\to$Cl &Rw$\to$Pr &Avg. & Sy$\to$Re\\
        \midrule
        Source        &-- &-- &\--  &43.7 &67.0 &73.9 &49.9 &60.1 &62.5 &51.7 &40.9 &72.6 &64.2 &46.3 &78.1 &59.2 &49.2 \\
        \midrule
        DAPL-RN~\cite{ge2022domain}       &TNNLS23 &\xmark  &\cmark   &54.1 &84.3 &84.8 &74.4 &83.7 &85.0 &74.5 &54.6 &84.8 &75.2 &54.7 &83.8 &74.5 &86.9 \\
        PADCLIP-RN~\cite{lai2023padclip}  &ICCV23  &\xmark  &\cmark   &57.5 &84.0 &83.8 &77.8 &85.5 &84.7 &76.3 &59.2 &85.4 &78.1 &60.2 &86.7 &76.6 &88.5 \\
        ADCLIP-RN~\cite{singha2023ad}    &ICCVW23  &\xmark  &\cmark   &55.4 &85.2 &85.6 &76.1 &85.8 &86.2 &76.7 &56.1 &85.4 &76.8 &56.1 &85.5 &75.9 &87.7 \\
        \midrule
        SHOT~\cite{liang2020we}      &ICML20  &\cmark &\xmark  &56.7 &77.9 &80.6 &68.0 &78.0 &79.4 &67.9 &54.5 &82.3 &74.2 &58.6 &84.5 &71.9 &82.7\\
        NRC~\cite{yang2021nrc}      &NIPS21  &\cmark  &\xmark  &57.7 &80.3 &82.0 &68.1 &79.8 &78.6 &65.3 &56.4 &83.0 &71.0 &58.6 &85.6 &72.2 &85.9 \\
        GKD~\cite{tang2021model}    &IROS21  &\cmark  &\xmark  &56.5 &78.2 &81.8 &68.7 &78.9 &79.1 &67.6 &54.8 &82.6 &74.4 &58.5 &84.8 &72.2 &83.0 \\ 
        
        AaD~\cite{yang2022attracting}      &NIPS22  &\cmark  &\xmark   &59.3 &79.3 &82.1 &68.9 &79.8 &79.5 &67.2 &57.4 &83.1 &72.1 &58.5 &85.4 &72.7 &88.0 \\
        AdaCon~\cite{chen2022contrastive}  &CVPR22  &\cmark  &\xmark   &47.2 &75.1 &75.5 &60.7 &73.3 &73.2 &60.2 &45.2 &76.6 &65.6 &48.3 &79.1 &65.0 &86.8 \\
        
        CoWA~\cite{lee2022confidence}      &ICML22  &\cmark  &\xmark  &56.9 &78.4 &81.0 &69.1 &80.0 &79.9 &67.7  &57.2 &82.4 &72.8 &60.5 &84.5 &72.5 &86.9 \\

        SCLM~\cite{tang2022sclm}      &NN22    &\cmark  &\xmark  &58.2 &80.3 &81.5 &69.3 &79.0 &80.7 &69.0 &56.8 &82.7 &74.7 &60.6 &85.0 &73.0 &85.3 \\
        ELR~\cite{yi2023source}       &ICLR23  &\cmark  &\xmark  &58.4 &78.7 &81.5 &69.2 &79.5 &79.3 &66.3 &58.0 &82.6 &73.4 &59.8 &85.1 &72.6 &85.8 \\
        PLUE~\cite{Litrico_2023_CVPR} &CVPR23  &\cmark  &\xmark  &49.1 &73.5 &78.2 &62.9 &73.5 &74.5 &62.2 &48.3 &78.6 &68.6 &51.8 &81.5 &66.9 &88.3 \\
        TPDS~\cite{tang2023source}    &IJCV23  &\cmark  &\xmark  &59.3 &80.3 &82.1 &70.6 &79.4 &80.9 &69.8 &56.8 &82.1 &74.5 &61.2 &85.3 &73.5 &87.6 \\
        \rowcolor{gray! 40} {\textbf{\modelshortname}-C-RN} &-- &\cmark &\cmark  &62.6 &87.5 &87.1 &79.5 &87.9 &87.4 &78.3 &63.4 &88.1 &80.0 &63.3 &87.7 &79.4 &88.8\\
        \rowcolor{gray! 40} {\textbf{\modelshortname}-C-B32} &-- &\cmark &\cmark   &\textbf{\color{cmred}70.6} &\textbf{\color{cmred}90.6} &\textbf{\color{cmred}88.8} &\textbf{\color{cmred}82.5} &\textbf{\color{cmred}90.6} &\textbf{\color{cmred}88.8} &\textbf{\color{cmred}80.9} &\textbf{\color{cmred}70.1} &\textbf{\color{cmred}88.9} &\textbf{\color{cmred}83.4} &\textbf{\color{cmred}70.5} &\textbf{\color{cmred}91.2} &\textbf{\color{cmred}83.1} 
        &\textbf{\color{cmred}90.3}\\
        \bottomrule
    \end{tabular}
\end{table*}

\begin{table*}[t]
    \caption{Closed-set SFDA on {\bf DomainNet-126} (\%). 
    \textbf{SF} and \textbf{M} means source-free and multimodal, respectively.
    }
    \label{tab:dn}
    \renewcommand\tabcolsep{5.5pt}
    \renewcommand\arraystretch{0.9}
    \scriptsize
    \centering
    \begin{tabular}{ l l|c c|c c c  c c c c c c c c c c }
        \toprule
        Method &{Venue} &{\bf SF} &{\bf M}
        &C$\to$P &C$\to$R &C$\to$S
        &P$\to$C &P$\to$R &P$\to$S    
        &R$\to$C &R$\to$P &R$\to$S  
        &S$\to$C &S$\to$P &S$\to$R &Avg.\\
        \midrule
        Source   &--   &--  &--  &44.6 &59.8 &47.5 &53.3 &75.3 &46.2 &55.3 &62.7 &46.4 &55.1 &50.7 &59.5 &54.7 \\
        \midrule
        DAPL-RN~\cite{ge2022domain} &TNNLS23  &\xmark &\cmark  &72.4 &87.6 &65.9 &72.7 &87.6 &65.6 &73.2 &72.4 &66.2 &73.8 &72.9 &87.8 &74.8 \\ 
        ADCLIP-RN~\cite{singha2023ad} &ICCVW23  &\xmark &\cmark  &71.7 &88.1 &66.0 &73.2 &86.9 &65.2 &73.6 &73.0 &68.4 &72.3 &74.2 &89.3 &75.2 \\ 
        \midrule
        SHOT~\cite{liang2020we}     &ICML20  &\cmark &\xmark  &63.5 &78.2 &59.5 &67.9 &81.3 &61.7 &67.7 &67.6 &57.8 &70.2 &64.0 &78.0 &68.1 \\
        GKD~\cite{tang2021model}    &IROS21  &\cmark &\xmark  &61.4 &77.4 &60.3 &69.6 &81.4 &63.2 &68.3 &68.4 &59.5 &71.5 &65.2 &77.6 &68.7 \\ 
        NRC~\cite{yang2021nrc}      &NIPS21  &\cmark &\xmark  &62.6 &77.1 &58.3 &62.9 &81.3 &60.7 &64.7 &69.4 &58.7 &69.4 &65.8 &78.7 &67.5 \\
        AdaCon~\cite{chen2022contrastive}    &CVPR22  &\cmark &\xmark &60.8 &74.8 &55.9 &62.2 &78.3 &58.2 &63.1 &68.1 &55.6 &67.1 &66.0 &75.4 &65.4 \\
        CoWA~\cite{lee2022confidence}        &ICML22  &\cmark &\xmark &64.6 &80.6 &60.6 &66.2 &79.8 &60.8 &69.0 &67.2 &60.0 &69.0 &65.8 &79.9 &68.6\\
        PLUE~\cite{Litrico_2023_CVPR}        &CVPR23  &\cmark &\xmark &59.8 &74.0 &56.0 &61.6 &78.5 &57.9 &61.6 &65.9 &53.8 &67.5 &64.3 &76.0 &64.7  \\
        TPDS~\cite{tang2023source}           &IJCV23  &\cmark &\xmark &62.9 &77.1 &59.8 &65.6 &79.0 &61.5 &66.4 &67.0 &58.2 &68.6 &64.3 &75.3 &67.1 \\
        \rowcolor{gray! 40}  {\textbf{\modelshortname}-C-RN} &-- &\cmark &\cmark &73.8 &\textbf{\color{cmred}89.0} &69.4 &74.0 &\textbf{\color{cmred}88.7} &70.1 &74.8 &74.6 &69.6 &74.7 &74.3 &\textbf{\color{cmred}88.0} &76.7 \\
        \rowcolor{gray! 40}  {\textbf{\modelshortname}-C-B32}  &--  &\cmark &\cmark 
        &\textbf{\color{cmred}76.6} &87.2 &\textbf{\color{cmred}74.9} &\textbf{\color{cmred}80.0} &87.4 &\textbf{\color{cmred}75.6} &\textbf{\color{cmred}80.8} &\textbf{\color{cmred}77.3} &\textbf{\color{cmred}75.5} &\textbf{\color{cmred}80.5} &\textbf{\color{cmred}76.7} &87.3 &\textbf{\color{cmred}80.0} \\
        \bottomrule
    \end{tabular}
\end{table*}

\begin{table*}[!tph]
    \caption{Results (\%) of CLIP and Source+CLIP on the four evaluation datasets. 
    The backbone of CLIP image-encoder in CLP-C-RN and CLP-C-B32 are the same as \textbf{\modelshortname}-C-RN and \textbf{\modelshortname}-C-B32, respectively. 
    The full results are provided in \texttt{Supplementary}.}
    \scriptsize
    \centering
    \label{tab:zeroshot}
    \setlength{\tabcolsep}{1.8mm}
    \renewcommand\arraystretch{0.9}
    \begin{tabular}{ll|cccc|ccccc|c|ccccccc}              
        \toprule                                          
        \multirow{2}{*}{Method}& \multirow{2}{*}{Venue}&  
        \multicolumn{4}{c}{\textbf{Office-31}}   \vline&  
        \multicolumn{5}{c}{\textbf{Office-Home}} \vline&  
        \multicolumn{1}{c}{\textbf{VisDA}}       \vline&
        \multicolumn{5}{c}{\textbf{DomainNet-126}}\cr     
        & &{$\to$A}  &{$\to$D}  &{$\to$W}    &{$\to$Avg.} &{$\to$Ar} &{$\to$Cl}
        &{$\to$Pr}   &{$\to$Rw} &{$\to$Avg.} &{Sy$\to$Re} &{$\to$C}  &{$\to$P} 
        &{$\to$R}    &{$\to$S}  &{$\to$Avg.} \\
        \midrule
        CLIP-RN~\cite{radford2021learning} &ICML21 &73.1   &73.9 &67.0 &71.4 &72.5 
         &51.9  &81.5  &82.5  &72.1 &83.7  &67.9 
        &70.2  &87.1 &65.4 &72.7 \\
        Source+CLIP-RN &-- &76.3 &90.4 &84.0 &83.6 &75.4 &57.4 &84.4 &85.7 &75.7 &82.0 &71.8 &71.4 &87.3 &66.5 &74.3 \\
        \rowcolor{gray! 40} \textbf{\modelshortname}-C-RN  &-- &\textbf{\color{cmred}78.6}&\textbf{\color{cmred}95.3}&\textbf{\color{cmred}93.9}&\textbf{\color{cmred}89.3} &\textbf{\color{cmred}79.3}&\textbf{\color{cmred}63.1}&\textbf{\color{cmred}87.7}&\textbf{\color{cmred}87.5}&\textbf{\color{cmred}79.4}
        &\textbf{\color{cmred}88.8}
        &\textbf{\color{cmred}74.5} &\textbf{\color{cmred}74.2}&\textbf{\color{cmred}88.5}&\textbf{\color{cmred}69.7}&\textbf{\color{cmred}76.7} \\
        \midrule
        CLIP-B32~\cite{radford2021learning} &ICML21 &76.0  &82.7 &80.6 &79.8 &74.6 &59.8 &84.3 &85.5 &76.1 &82.9 &74.7 &73.5 &85.7 &71.2 &76.3 \\
        Source+CLIP-B32 &-- &78.5 &93.0 &89.6 &87.0 &78.9 &62.5 &86.1 &87.7 &78.8 &82.0 &76.8 &73.7 &86.0 &70.8 &76.8 \\
        \rowcolor{gray! 40} \textbf{\modelshortname}-C-B32  &-- &\textbf{\color{cmred}83.1}&\textbf{\color{cmred}98.0}&\textbf{\color{cmred}96.4}&\textbf{\color{cmred}92.5} &\textbf{\color{cmred}82.3}&\textbf{\color{cmred}70.4}&\textbf{\color{cmred}90.8}&\textbf{\color{cmred}88.8}&\textbf{\color{cmred}83.1}
        &\textbf{\color{cmred}90.3}
        &\textbf{\color{cmred}80.4} &\textbf{\color{cmred}76.9}&\textbf{\color{cmred}87.3}&\textbf{\color{cmred}75.3}&\textbf{\color{cmred}80.0} \\
        \bottomrule
    \end{tabular}
\end{table*}


\paragraph{Datasets.} 
We evaluate four standard benchmarks:  
\textbf{Office-31}~\cite{saenko2010adapting}, 
\textbf{Office-Home}~\cite{venkateswara2017deep}, 
\textbf{VisDA}~\cite{peng2017visda} 
and \textbf{DomainNet-126}~\cite{peng2019moment}. 
Among them, \textbf{Office-31} is a small-scaled dataset; \textbf{Office-Home} is a medium-scale dataset; \textbf{VisDA} and \textbf{DomainNet-126} are both large-scale dataset.
The details of the four datasets are provided in \texttt{Supplementary}.


\vspace{4pt}
\noindent\textbf{Competitor.}
We compare {\modelshortname} with 18 existing top-performing methods into three groups. 
(1) {\itshape The first group} contains Source (the source model's results), CLIP~\cite{radford2021learning} and Source+CLIP where Source+CLIP directly average the results of the source model and CLIP. 
(2) {\itshape The second group} includes three state-of-the-art {UDA} methods DAPL~\cite{ge2022domain}, PADCLIP~\cite{lai2023padclip} and ADCLIP~\cite{singha2023ad} that are also multimodal guiding-based. 
(3) {\itshape The third group} comprises 13 current state-of-the-art {SFDA} models: 
SHOT~\cite{liang2020we}, 
NRC~\cite{yang2021nrc},
GKD~\cite{tang2021model}, 
HCL~\cite{huang2021model},
AaD~\cite{yang2022attracting},
AdaCon~\cite{chen2022contrastive},
CoWA~\cite{lee2022confidence},
SCLM~\cite{tang2022sclm}, 
ELR~\cite{yi2023source},
PLUE~\cite{Litrico_2023_CVPR}, 
TPDS~\cite{tang2023source} and
CRS~\cite{zhang2023class}.


For comprehensive comparisons, we implement {\modelshortname} in two variants: (1) {\modelshortname}-C-RN (weak version) and (2) {\modelshortname}-C-B32 (strong version). The key distinction lies in the backbone of the CLIP image-encoder. Specifically, for {\modelshortname}-C-RN, ResNet101~\cite{he2016deep} is employed on the VisDA dataset, while ResNet50~\cite{he2016deep} is used on the other three datasets. On the other hand, {\modelshortname}-C-B32 adopts ViT-B/32~\cite{han2022survey} as the backbone across all datasets.

\vspace{4pt}
\noindent\textbf{SFDA settings.}
We consider three distinct settings: 
the conventional closed-set SFDA setting,  the partial-set and 
the open-set SFDA settings. 
The experiment implementation details are provided in \texttt{Supplementary}.


\subsection{Comparison Results}
\textbf{Comparison on Closed-set SFDA setting.}
The comparisons of the four evaluation datasets are listed in Tab.~\ref{tab:oc}$\sim$\ref{tab:dn}. 
{\modelshortname}-C-B32 surpasses the previous best method CoWA~(on Office-31), TPDS~(on Office-Home) and PLUE~(on VisDA) and GKD~(on DomainNet-126) by \textbf{2.2\%}, \textbf{9.6\%} \textbf{2.0\%} and \textbf{11.3\%} in average accuracy respectively.
Specifically, {\modelshortname}-C-B32 obtains the best results on 4 out of 6 tasks on {Office-31} while surpassing previous methods on all tasks of the other three datasets.
As for {\modelshortname}-C-RN, besides Office-31, it obtains the second-best results and beat the previous best methods by {\bf 5.9}\%, {\bf 0.5}\% and {\bf 8.0}\% on Office-Home, VisDA and DomainNet-126 in average accuracy.  
The comparison of {\modelshortname}-C-RN shows that our method can still perform well despite using a weaker CLIP.  
Based on a strong CLIP (see results of {\modelshortname}-C-B32), our method's performance can improve further as we expected. 
All of the results indicate that the {\modelshortname} can boost the cross-domain 
performance in closed-set SFDA setting. 

\vspace{3pt}
\noindent\textbf{Comparison to CLIP based prediction results.} 
The original CLIP model can conduct general image classification.  
We carry out a quantitative comparison between \modelshortname's adaptation performance and CLIP's performance on the four datasets, averaging the adaptation results of {\modelshortname} grouped by the target domain name.

As presented in the bottom of Tab.~\ref{tab:zeroshot}, {\modelshortname}-C-B32 outperforms CLIP-B32 on all tasks. 
On average accuracy, {\modelshortname}-C-B32 increases the performance by {\bf 12.7}\%, {\bf 7.0}\%, {\bf 7.4}\% and {\bf 3.7}\% in Office-31, Office-Home, VisDA and DomainNet-126, respectively. 
Regarding the weak version, as reported in the top, {\modelshortname}-C-RN maintains similar advantages with the increase of {\bf 17.9}\%, {\bf 7.3}\%, {\bf 5.1}\% and {\bf 4.0}\%.   
The result shows that \textit{the domain generality of the original CLIP model cannot fully excel to the target domain, and task-specific customization is needed.}

\begin{figure}[t]
    \centering
    \setlength{\abovecaptionskip}{0.15cm}
    \setlength{\belowcaptionskip}{-0.3cm}
    \includegraphics[width=0.95\linewidth]{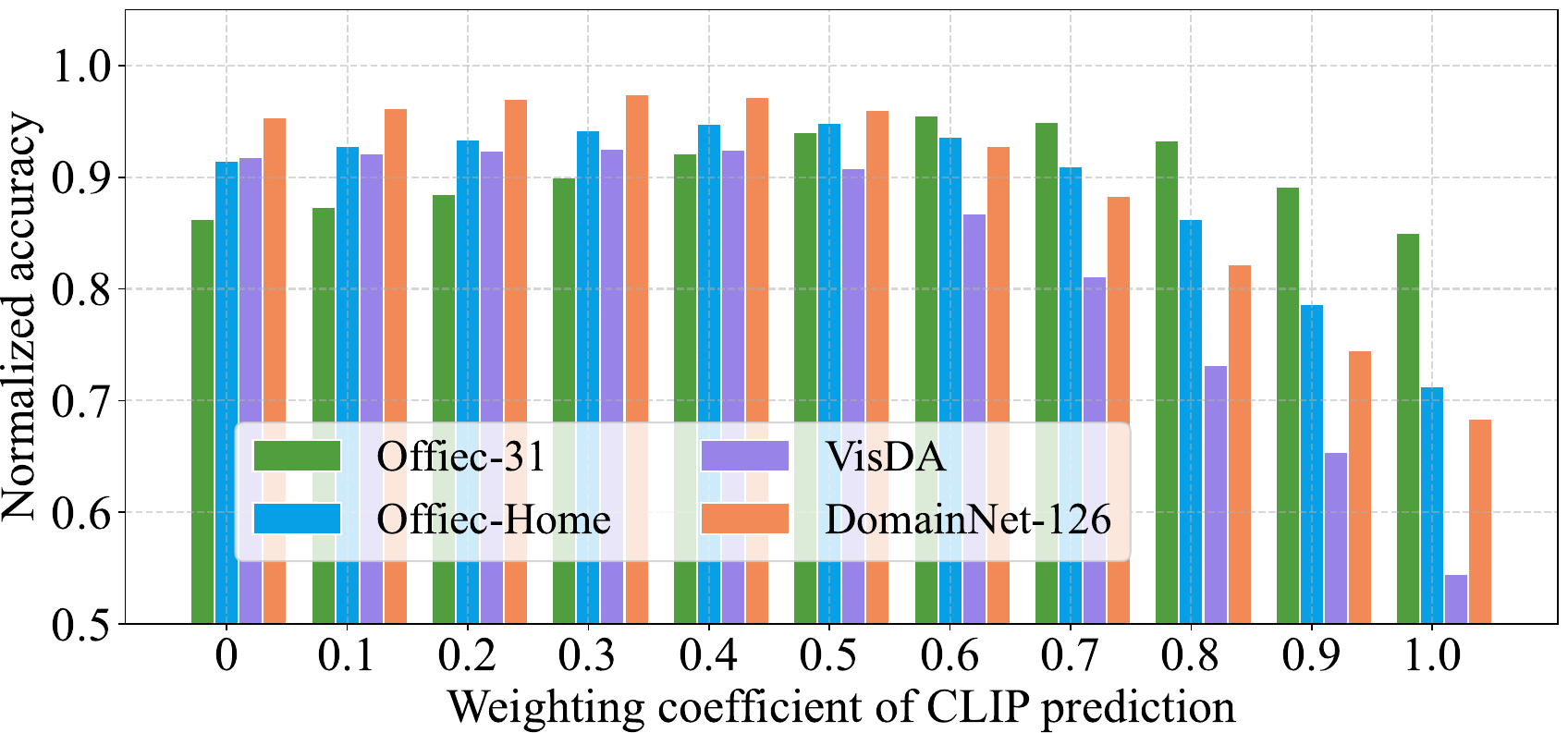}
    \caption{
     The performance of the scheme directly weighting the source model and CLIP-B32. 
     All results are normalized by corresponding {\modelshortname}-C-B32 accuracies for a clear view.
    } 
    \label{fig:sc-weight}
\end{figure}

Interestingly, compared with CLIP-B32, except for VisDA with a tiny gap of {\bf 0.9}\%, Source+CLIP-B32 averagely improve by {\bf 7.2}\%  at most on the other datasets. 
Meanwhile, Source+CLIP-B32 is beaten by {\modelshortname}-C-B32 with an increase of {\bf 3.2}\% at least. 
In the group of {\modelshortname}-C-RN, we have the same observation.   
These results imply that directly weighting the source model and CLIP is an intuitive knowledge adaptation scheme, but it is hard to perform adaptation deeply. 
Considering Source+CLIP is an average version, we conduct a comprehensive comparison with the weighting strategy where the weighting coefficient of CLIP prediction varies from 0.0 to 1.0. 
Here, we conduct this experiment based on more challenging CLIP-B32 due to its large performance gap with Source (see the first row in \texttt{Tab.\ref{tab:oc}$\sim$\ref{tab:dn}}).  
For a clear view, all weighted accuracies are normalized by the corresponding {\modelshortname}-C-B32 accuracies, respectively.
As shown in Fig.~\ref{fig:sc-weight}, no result can exceed the value of 1.0.
This indicates that \textit{weighting the source model and CLIP in a zero-shot manner cannot obtain desirable task-specific fusion, and a carefully designed distilling is necessary.}



\vspace{3pt}
\noindent\textbf{Comparison on Partial-set and Open-set SFDA settings.}
These are the variations of traditional Closed-set SFDA setting, following the same as SHOT~\cite{liang2020we} (the detailed setting introduction is provided in \texttt{Supplementary}). 
As reported in Tab.~\ref{tab:ps-os}, compared with previous best method CoWA~(Partial-set) and CRS~(Open-set), our {\modelshortname}-C-B32 improves by {\bf 2.4}\% and {\bf 2.7}\%, respectively. 

\subsection{Model Analysis} 

\begin{figure*}[t]
    \centering
    {\includegraphics[width=0.15\linewidth,height=0.155\linewidth]{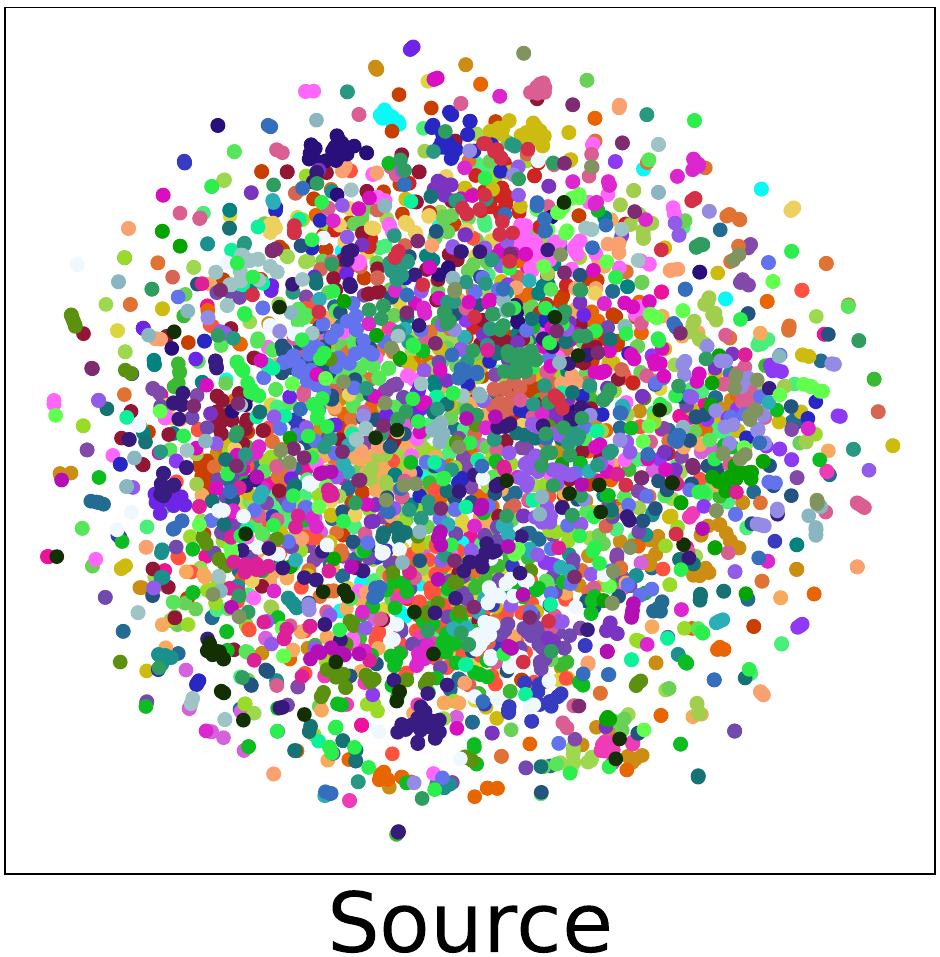}}
    {\includegraphics[width=0.15\linewidth,height=0.155\linewidth]{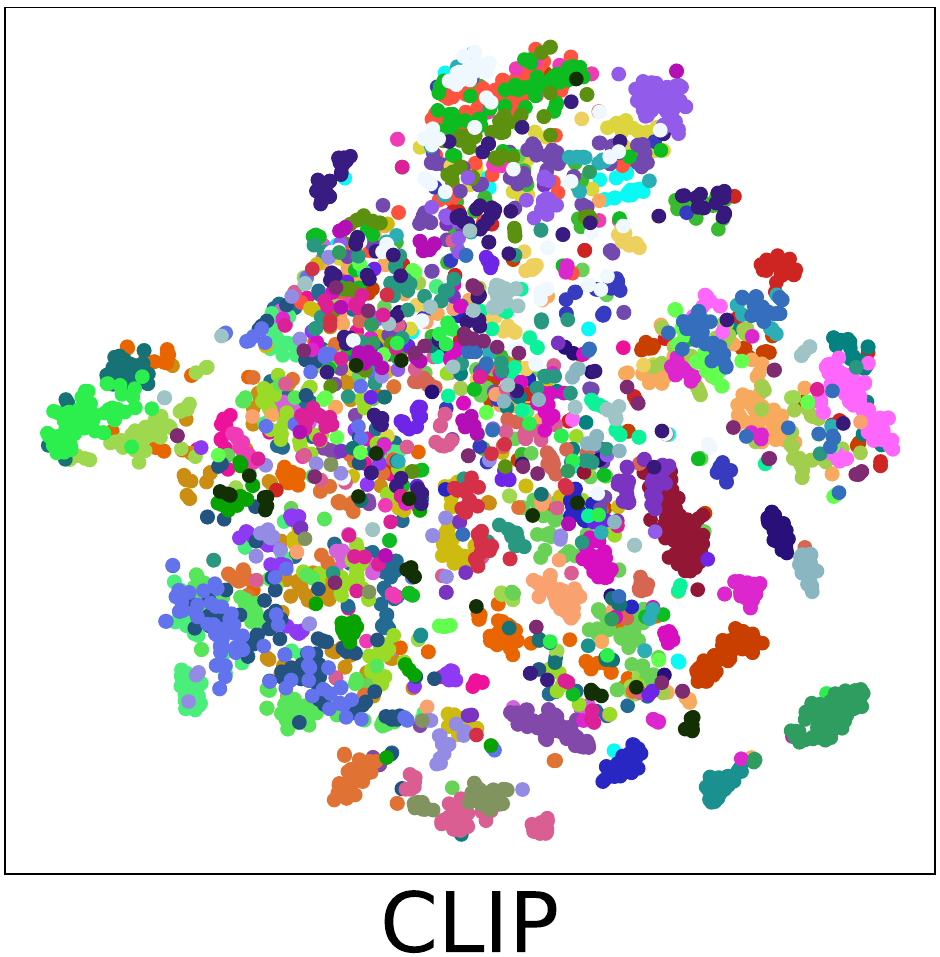}}
    {\includegraphics[width=0.15\linewidth,height=0.155\linewidth]{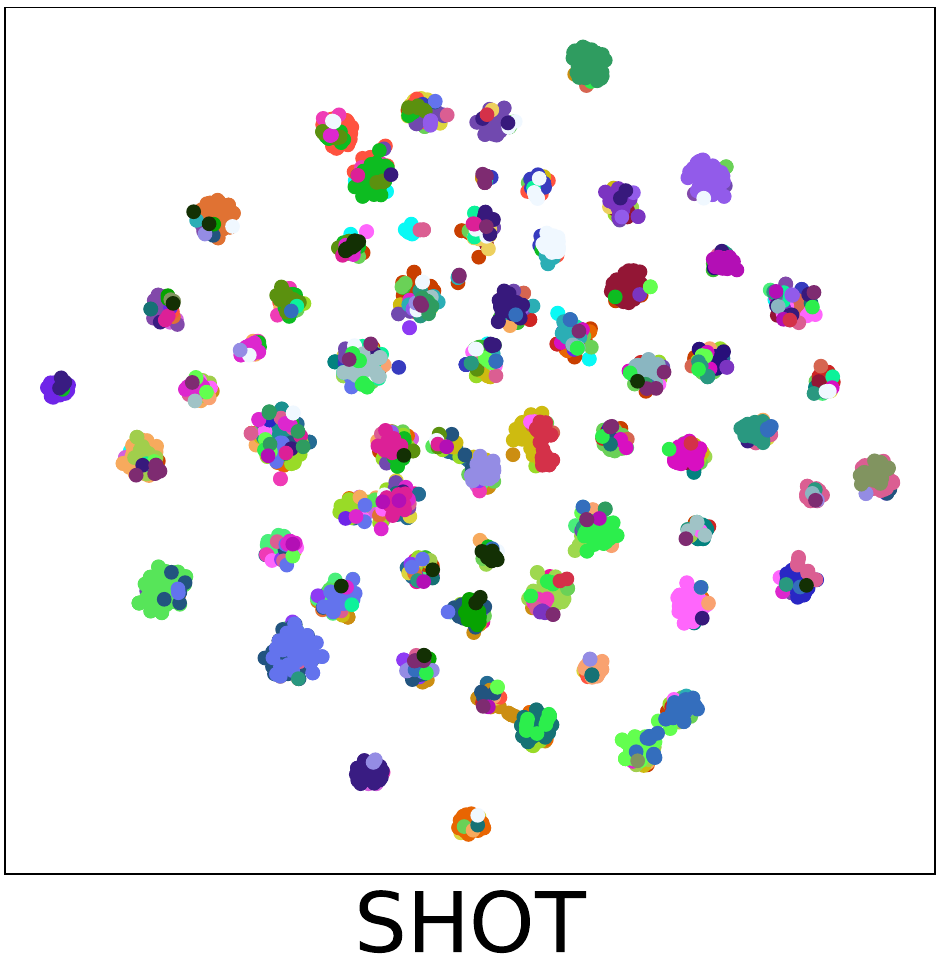}}
    {\includegraphics[width=0.15\linewidth,height=0.155\linewidth]{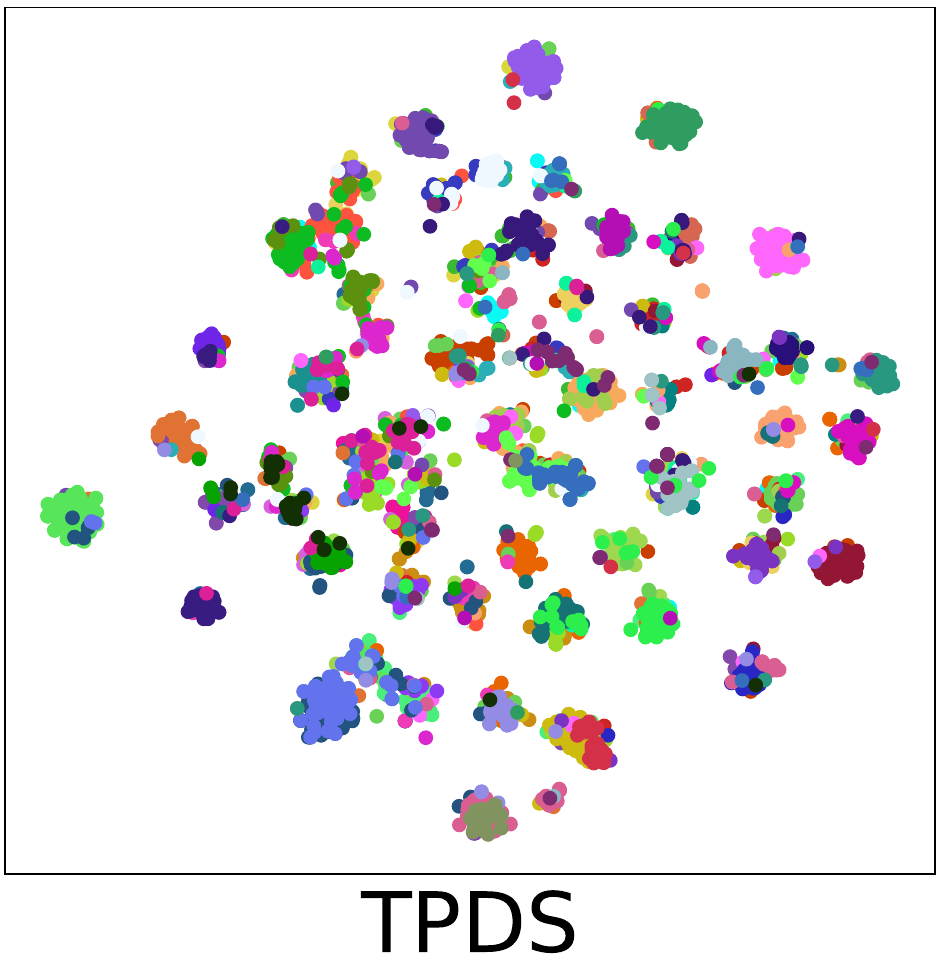}}
    {\includegraphics[width=0.15\linewidth,height=0.155\linewidth]{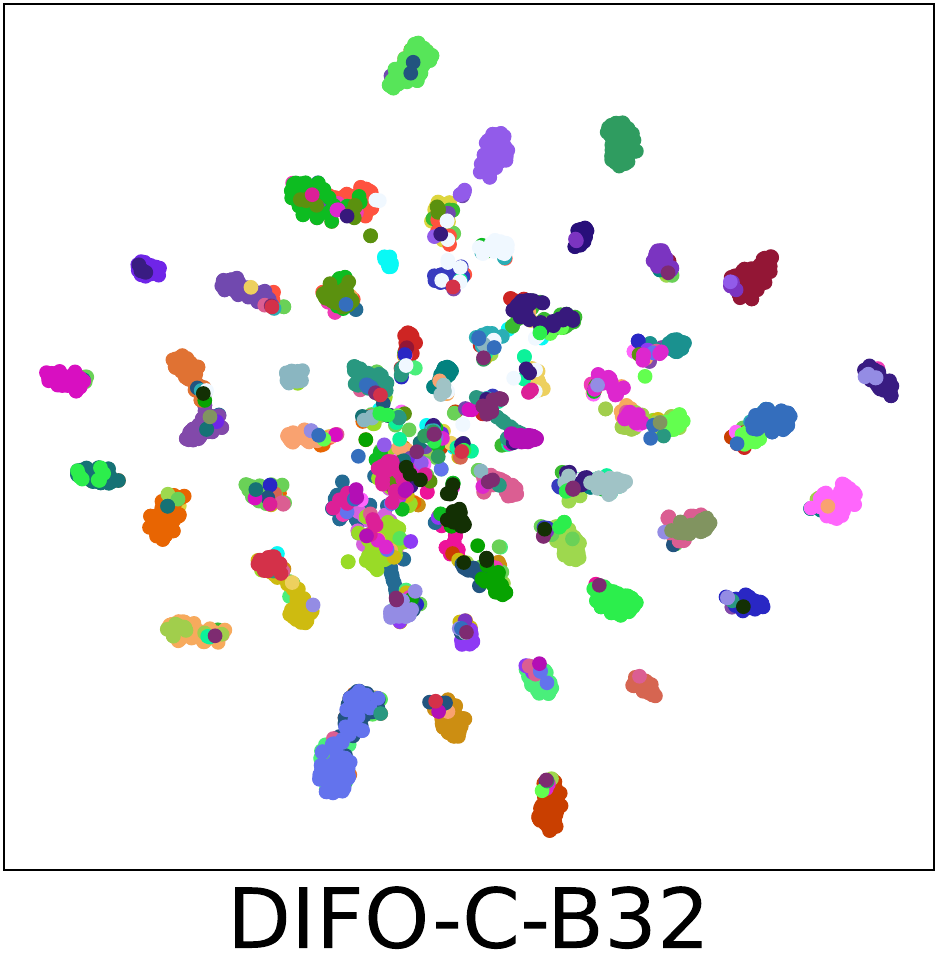}}
    {\includegraphics[width=0.15\linewidth,height=0.155\linewidth]{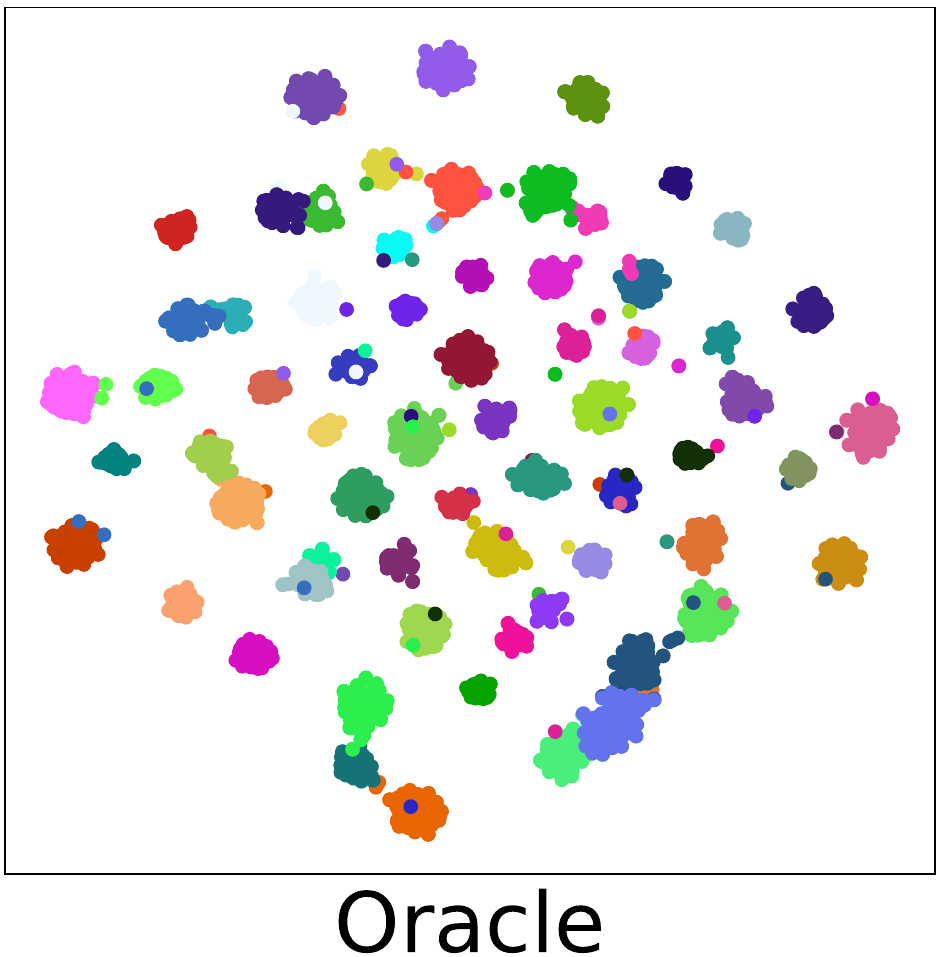}}\\
    \vspace{5pt}
    {\includegraphics[width=0.15\linewidth,height=0.155\linewidth]{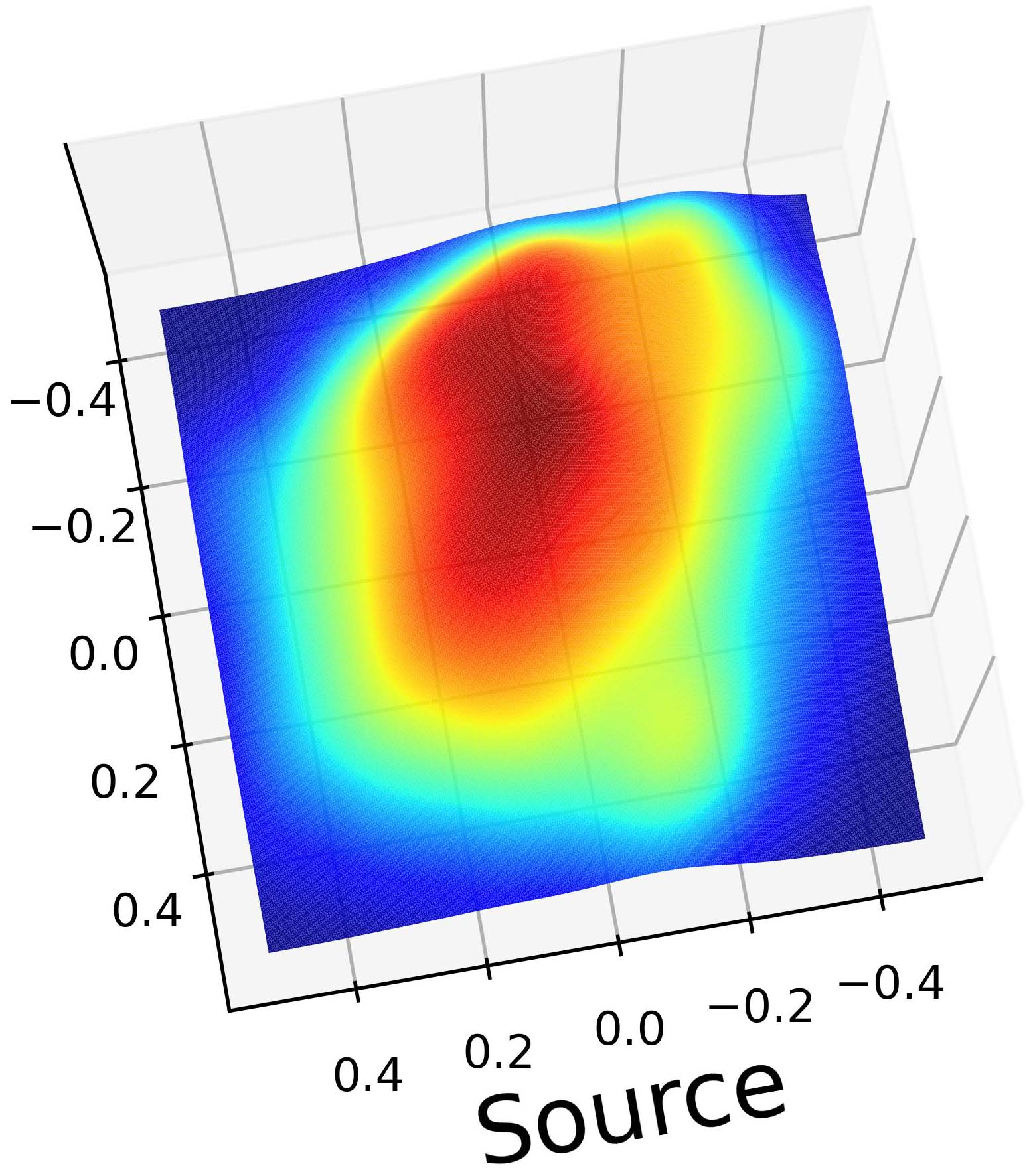}}
    {\includegraphics[width=0.15\linewidth,height=0.155\linewidth]{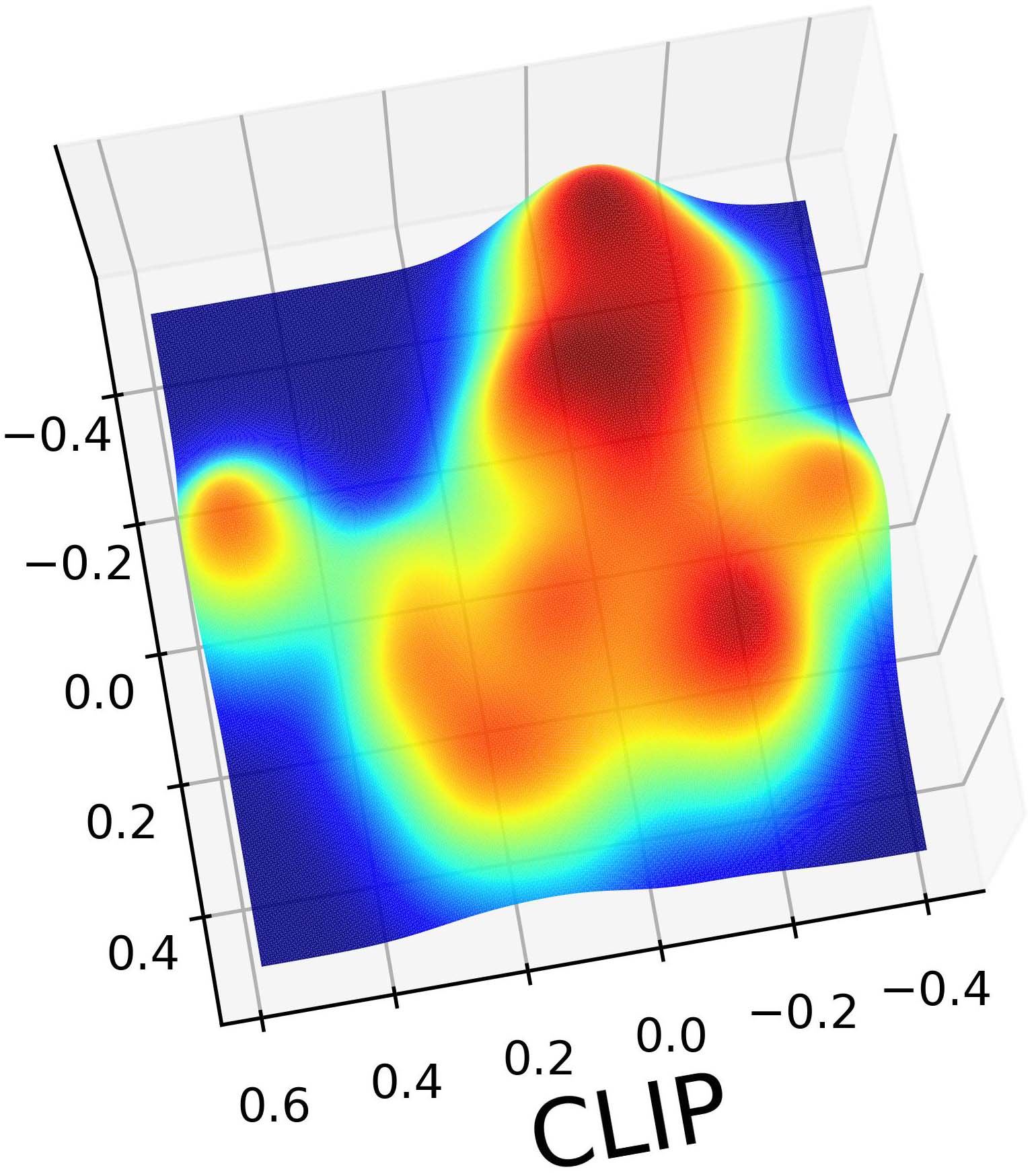}}
    {\includegraphics[width=0.15\linewidth,height=0.155\linewidth]{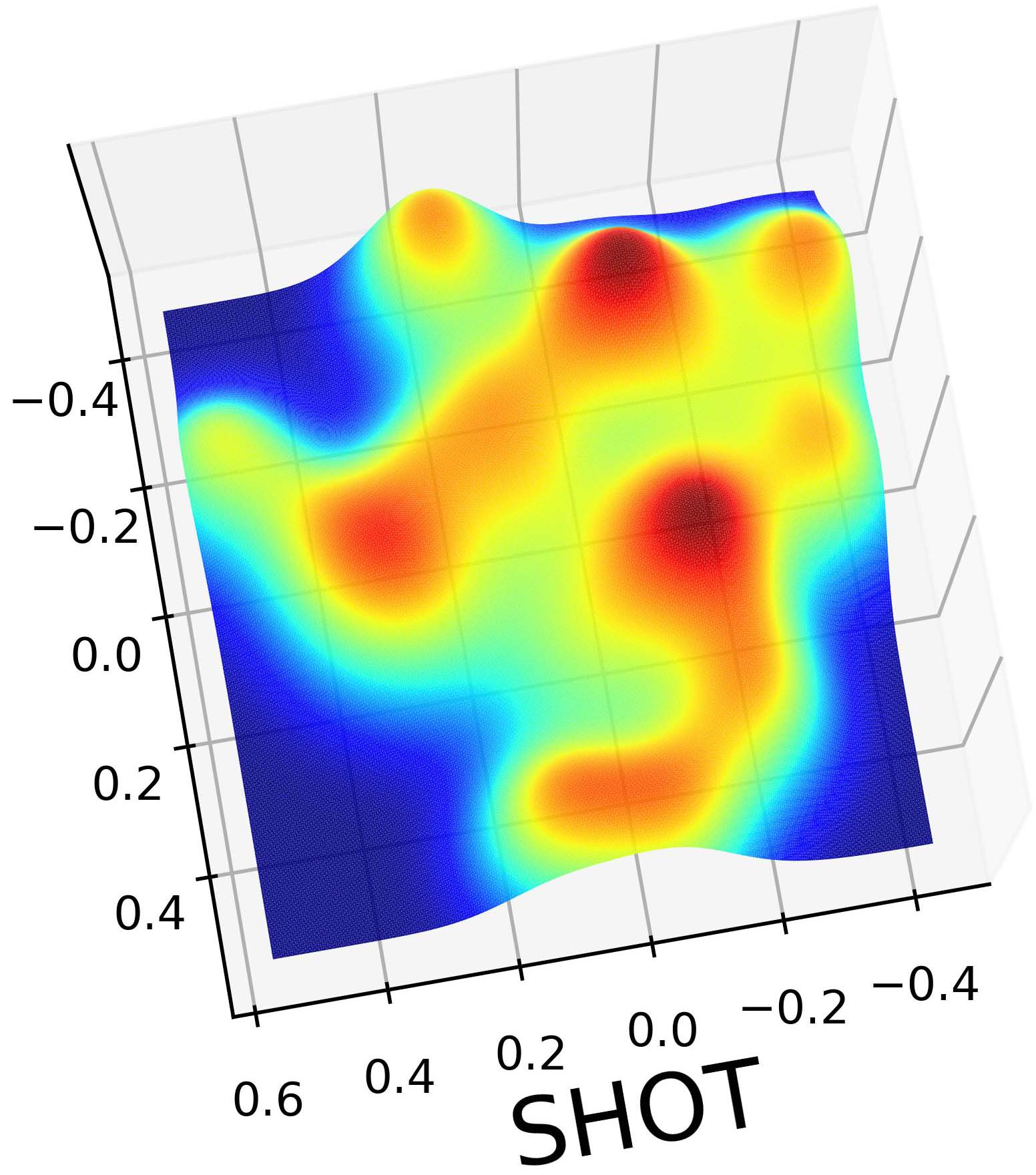}}
    {\includegraphics[width=0.15\linewidth,height=0.155\linewidth]{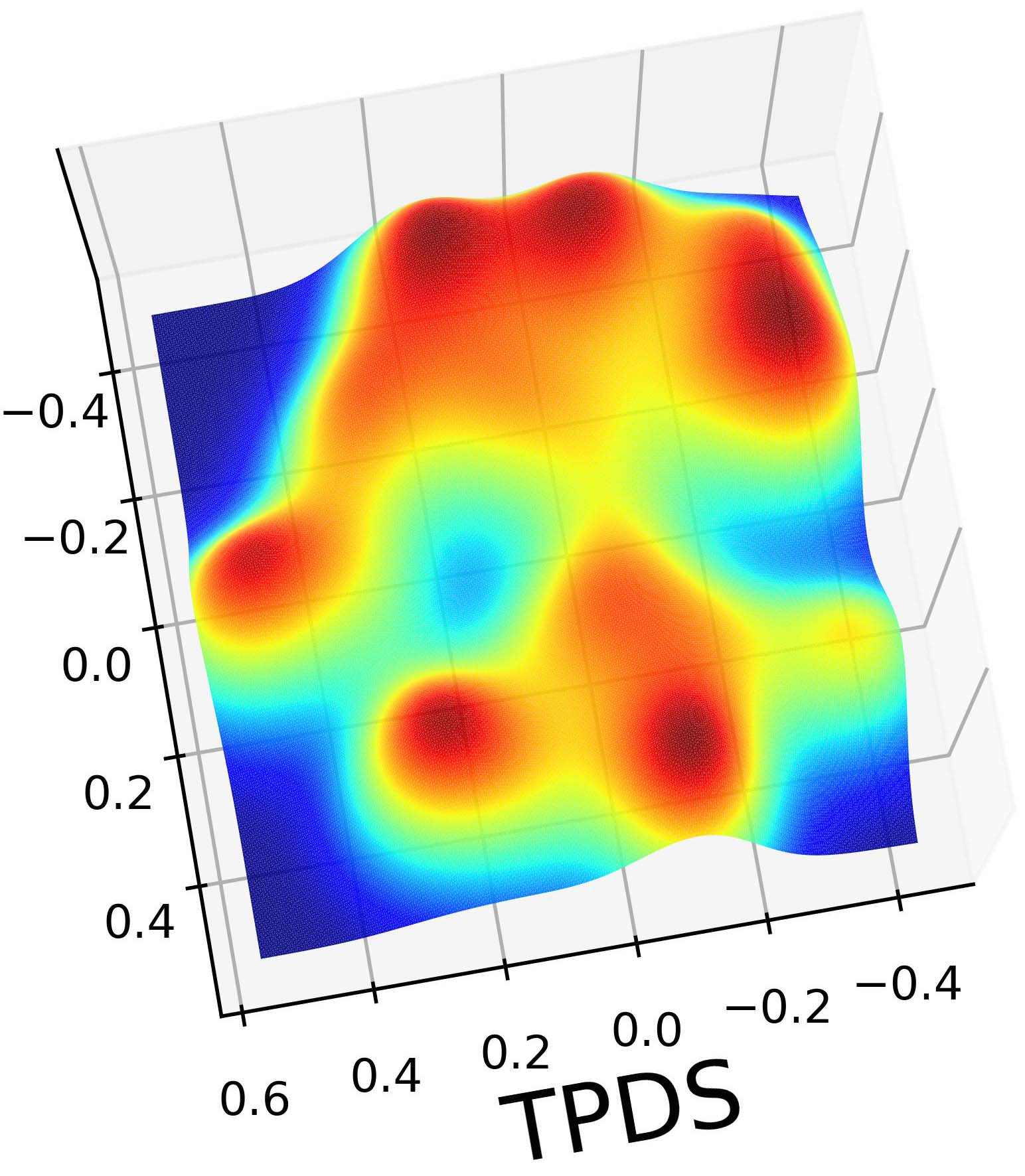}}
    {\includegraphics[width=0.15\linewidth,height=0.155\linewidth]{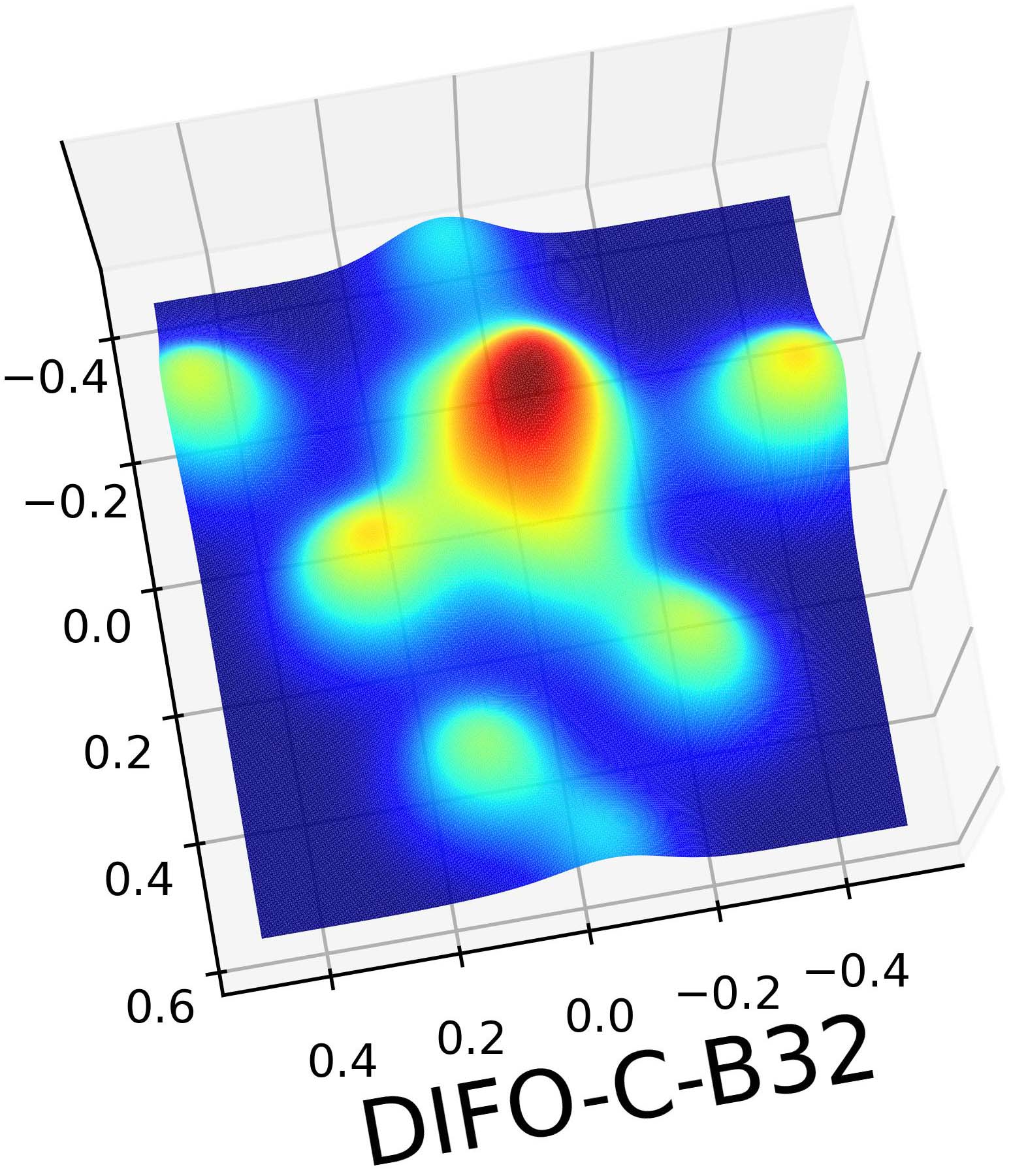}}
    {\includegraphics[width=0.15\linewidth,height=0.155\linewidth]{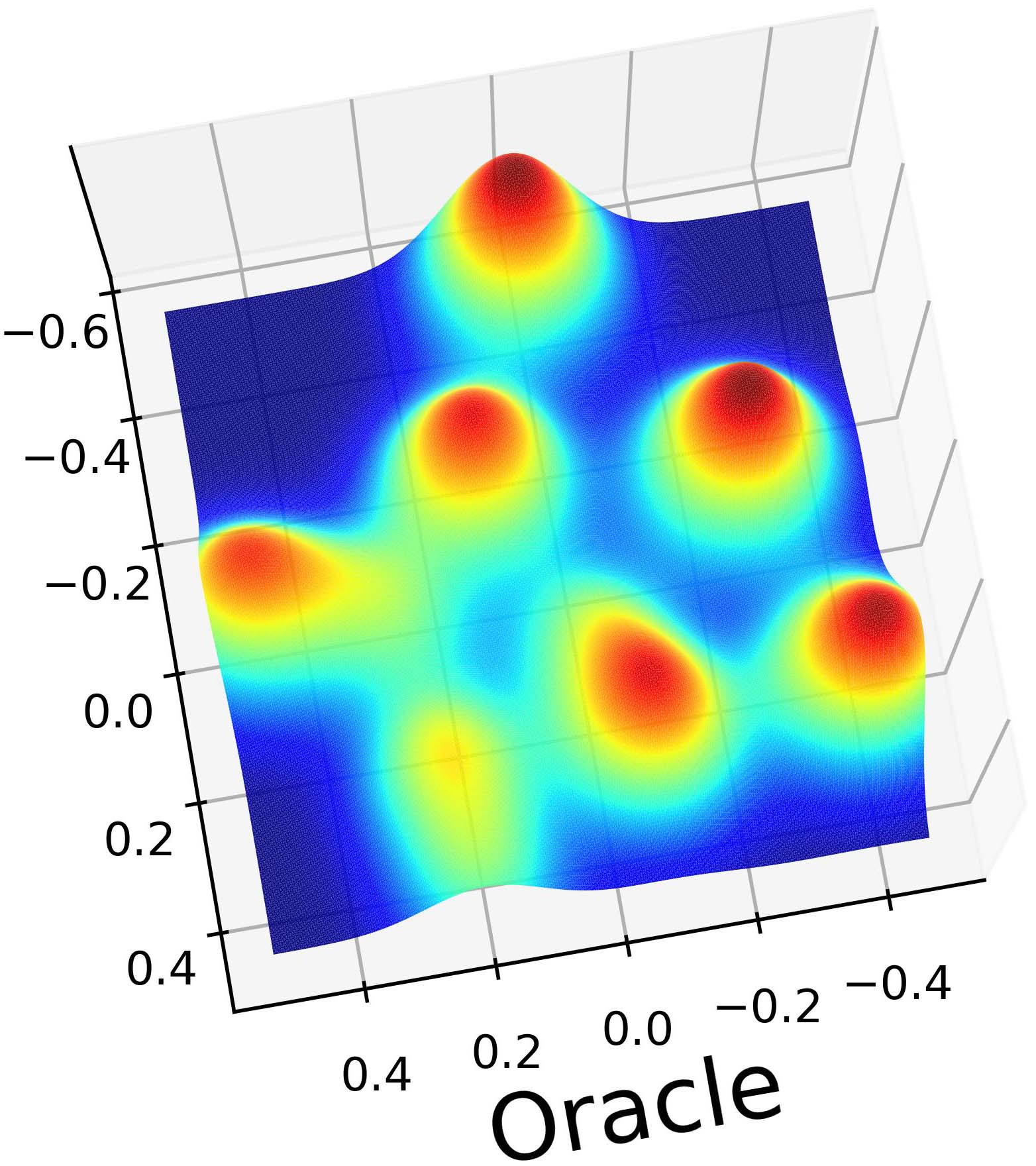}}
    \setlength{\abovecaptionskip}{0.3cm}
    \caption{
     Feature distribution visualization comparison on transfer task Ar$\to$Cl in Office-Home. 
     Oracle is trained on target domain Cl using the ground-truth labels. 
     Different colors stand for different categories. 
     {\bf Top}: t-SNE feature distribution over 65 categories.  
     {\bf Bottom}: The corresponding 3D density charts. 
     For easy view, the first 10 categories were used in this plot. 
     } 
    \label{fig:visfea}
\end{figure*}

\textbf{Feature distribution visualization.} 
Taking task Ar$\to$Cl in Office-Home as a toy experiment, we visualize feature distribution using t-SNE tool.
Meanwhile, we choose 5 comparisons, including the source model (termed Source), CLIP-B32’s zero shot (termed CLIP), SHOT, TPDS and Oracle (trained on domain Cl with the real labels). 
As shown at the top of Fig.~\ref{fig:visfea}, from Souce to {\modelshortname}-C-B32, category aliasing gradually relieves. 
Compared with Oracle, {\modelshortname}-C-B32 has the most similar distribution shape. 
To verify this point, we also give the 3D Density chart results arranged at the bottom of Fig.~\ref{fig:visfea}. 
These results confirm the effectiveness of our {\modelshortname}-C-B32 in terms of Feature distribution.

\vspace{4pt}
\noindent\textbf{Ablataion study.}
We evaluate the (1) effect of $L_{\rm{TSC}}$, $L_{\rm{MCE}}$ and $L_{\rm{PC}}$, (2) effect of optimization of mutual information, (3) effect of task-specific customization and (4) effect of historical prediction fusion.

For this first issue, we conduct a progressive experiment to isolate the loss's effects.
The top four rows of Tab.~\ref{tab:ab_loss} list the ablation study results. 
For convenience comparison, the baseline~(the first row) is the source model results.
When single $L_{\rm{TSP}}$ or $L_{\rm{MCE}}$ works, the accuracy largely increases on the three datasets with an improvement of about {\bf 20}\% in average accuracy compared with the baseline. 
As both of them are adopted, the accuracy evident increase ({\bf 3.7}\% in average, the fourth row) on the top of the case of only $L_{\rm{TSC}}$ and further enhanced by adopting of item $L_{\rm{PC}}$~({\bf 3.5}\% in average, the fifth row). 
The results indicate: (1) all objective components positively affect the final performance, (2) $L_{\rm{MCE}}$, $L_{\rm{PC}}$ is crucial due to providing a new soft supervision for coarse-to-fine adaptation.

For the second and third issues, we propose two variation methods of {\modelshortname}-C-B32 to evaluate the effect. 
One is {\modelshortname}-C-B32 w/ ${\rm{KL}}$ where the mutual information maximization loss in $L_{\rm{TSC}}$, $L_{\rm{PC}}$ are replace by KL divergence loss.
The other one is {\modelshortname}-C-B32 w/ CLIP where the prompt learning-based customization for CLIP is cancelled, and the inputted prompt is set to the fixed template of \textit{"a photo of a [CLS]."} during the entire adaptation.  
As presented in the last two rows in Tab.~\ref{tab:ab_loss}, {\modelshortname}-C-B32 (the fifth row) beats {\modelshortname}-C-B32 w/ ${\rm{KL}}$ and  {\modelshortname}-C-B32 w/ CLIP with average improvement of {\bf 1.6}\% at least, respectively confirming the effect of adopting mutual information optimization and task-specific customization. 
As for the fourth issue, its effect is verified by the performance decreases ({\bf 3.4}\% in average at most) in the variation methods (the last two rows), which remove $\boldsymbol{p}'_{i}$ and $\boldsymbol{p}_{i}$ from the fusion respectively. 


\begin{table}[t]
    \centering
    \renewcommand\tabcolsep{4.5pt}
    \renewcommand\arraystretch{0.9}
    \scriptsize
    \caption{{Partial-set SFDA} and {Open-set SFDA} on {\bf Office-Home} (\%).
              The full results are provided in \texttt{Supplementary}.
              }
        \begin{tabular}{llc|llc}
        \toprule
        Partial-set SFDA &Venue &{Avg.} &Open-set SFDA  &Venue &{Avg.} \\
        \midrule
        Source  &--  &{62.8} &Source 
        &--&{46.6} \\
        \midrule
        SHOT~\cite{liang2020we}         &ICML20  &79.3   &SHOT~\cite{liang2020we}        &ICML20 &{72.8} \\
        HCL~\cite{huang2021model}       &NIPS21  &79.6   &HCL~\cite{huang2021model}      &NIPS21 &{72.6} \\
        CoWA~\cite{lee2022confidence}   &ICML22  &83.2   &CoWA~\cite{lee2022confidence}  &ICML22 &{73.2} \\
        AaD~\cite{yang2022attracting}   &NIPS22  &79.7   &AaD~\cite{yang2022attracting}  &NIPS22 &{71.8} \\
        CRS~\cite{zhang2023class}       &CVPR23  &80.6   &CRS~\cite{zhang2023class}      &CVPR23 &{73.2} \\
        \rowcolor{gray! 40} \textbf{\modelshortname}-C-B32 &--  &\textbf{\color{cmred}85.6} &\textbf{\modelshortname}-C-B32 
        &--&\textbf{\color{cmred}75.9}  \\
        \midrule
    \end{tabular} 
    \label{tab:ps-os}
\end{table}

\begin{table}[t]
    \centering
    \renewcommand\tabcolsep{4.5pt}
    \renewcommand\arraystretch{0.9}
    \scriptsize
    \caption{Classification results of ablation study~(\%) on {\bf Office-31} {\bf Office-Home} and {\bf VisDA}.}
        \begin{tabular}{ccc|ccc|c}
        \toprule
        {$L_{\rm{TSC}}$} &{$L_{\rm{MCE}}$} &{$L_{\rm{PC}}$} &{\bf Office-31} &{\bf Office-Home} &{\bf VisDA} &{Avg.} \\
        \midrule
        \xmark  &\xmark  &\xmark  &{78.6} &{59.2} &{49.2} &{62.3} \\
        \cmark  &\xmark  &\xmark  &{82.4} &{77.4} &{84.4} &{81.4}\\
        \xmark  &\cmark  &\xmark  &{82.1} &{76.5} &{88.6} &{82.4} \\
        \cmark  &\cmark  &\xmark  &{87.0} &{80.0} &{88.3} &{85.1} \\
        \cmark  &\cmark  &\cmark  &\textbf{\color{cmred}92.5} &\textbf{\color{cmred}83.1} &\textbf{\color{cmred}90.3}
        &\textbf{\color{cmred}88.6} \\
        \midrule
        \multicolumn{3}{l}{\textbf{\modelshortname}-C-B32 w/ ${\rm{KL}}$} \vline &{90.4} &{81.5} &{89.0} &{87.0} \\
        \multicolumn{3}{l}{\textbf{\modelshortname}-C-B32 w/ CLIP}     \vline &{90.7} &{81.1} &{88.8} &{86.8} \\
        \multicolumn{3}{l}{\textbf{\modelshortname}-C-B32 w/o $\boldsymbol{p}'_{i}$} \vline  &{89.8} &{73.5} &{87.0} &{83.4} \\
        \multicolumn{3}{l}{\textbf{\modelshortname}-C-B32 w/o ${\boldsymbol{p}}_{i}$} \vline &{88.9} &{82.2} &{88.9} &{86.7} \\
        \midrule
    \end{tabular} 
    \label{tab:ab_loss}
\end{table}



\begin{figure}[t]
    \centering
    {\includegraphics[width=0.49\linewidth,height=0.47\linewidth]{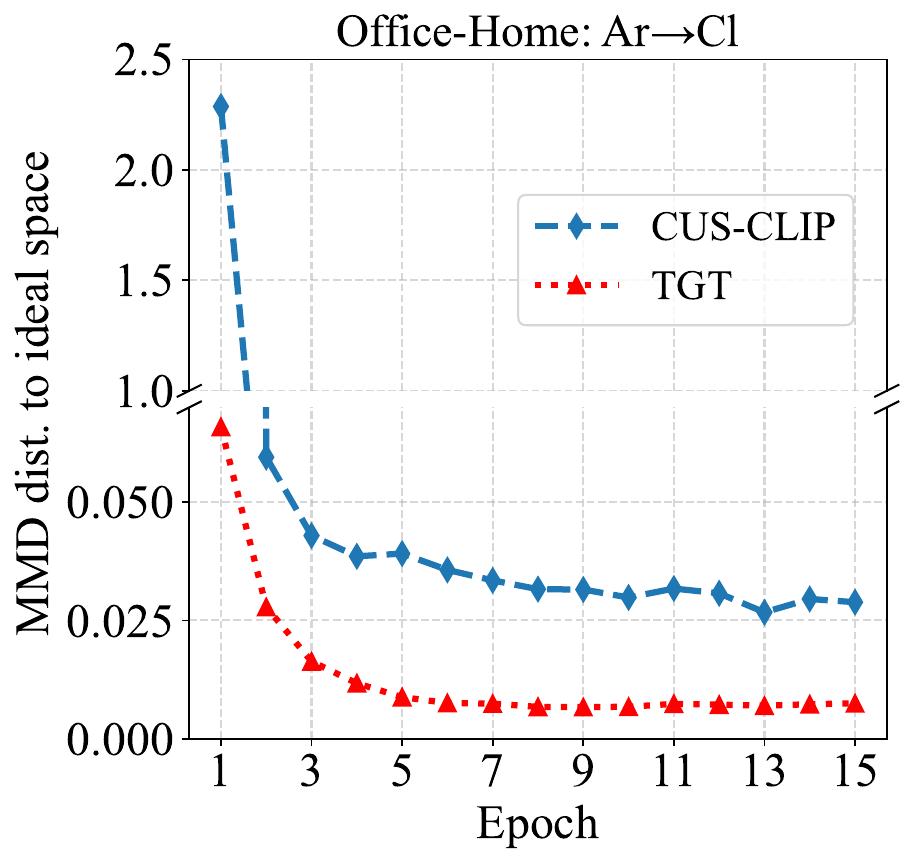}}
    {\includegraphics[width=0.49\linewidth,height=0.47\linewidth]{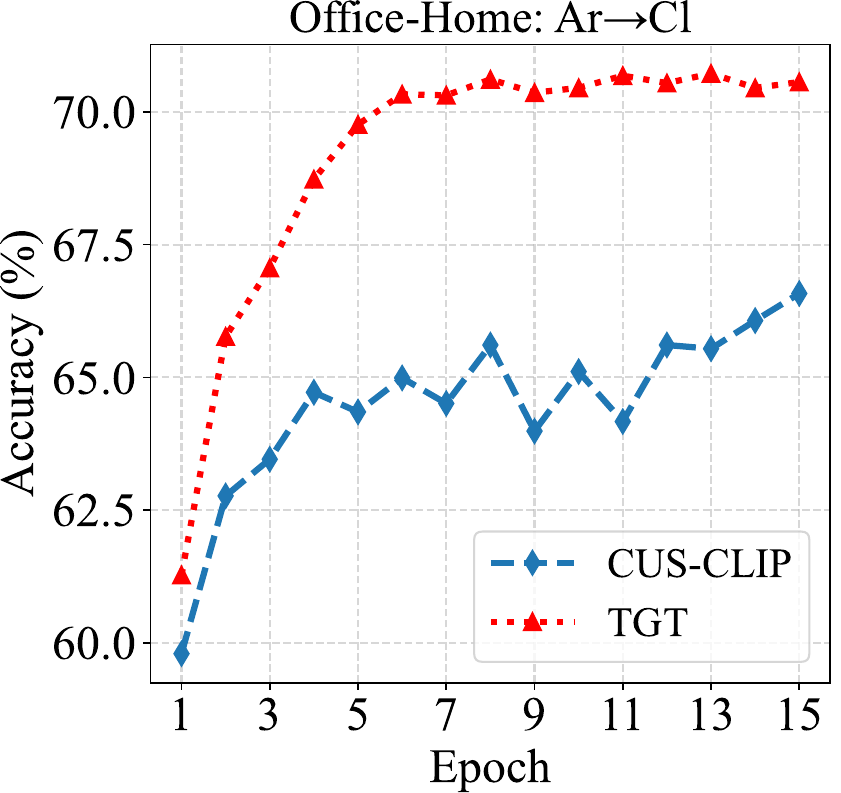}}
    \caption{
     The evolving dynamics of MMD distance during adaption of Ar$\to$Cl in Office-Home. 
     {\bf Left} and {\bf Right} present the varying curves of MMD distance and accuracy, respectively
    } 
    \label{fig:dis-analy}
\end{figure}

\subsection{Task-Specific Knowledge Adaptation Analysis} \label{sec:tskaa}
In this part, we give a feature space shift analysis using the measure of MMD (maximum mean
discrepancy) distance~\cite{MMDLoss2019} to verify whether the proposed method ensures a task-specific knowledge adaptation.

In this experiment, we first train a domain-invariant Oracle model over all Office-Home data with real labels, and use the logits to express the ideal task-specific space $\mathcal{O}$. 
After that, an analysis is conducted on the transfer task Ar$\to$Cl. 
During this adaptation, there are $T$ (epoch number) intermediate target models and customized CLIP models. 
We feedforward the target data through each intermediate model and take the logits as a space. 
Thus, we obtain $T$ intermediate target feature spaces $\{\mathcal{U}_k\}_{k=1}^{T}$ and $T$ intermediate customized CLIP feature spaces $\{\mathcal{V}\}_{k=1}^{T}$. 
Within this context, these intermediate spaces can depict the task-specific distillation to $\mathcal{O}$.
In practice, the CLIP image encoder's backbone is set to ViT-B/32.

In the left of Fig.~\ref{fig:dis-analy}, we give the MMD distance change curve of $\{\mathcal{U}_k\}_{k=1}^{T}$ (in red, termed {TGT}) and $\{\mathcal{V}\}_{k=1}^{T}$ (in blue, termed CUS-CLIP), taking $\mathcal{O}$ as the original space. 
It is seen that at early epochs (1$\sim$4), TGT and CUS-CLIP sharply decrease and then maintain a gradual decrease in the following epochs. 
Meanwhile, this change is consistent with the accuracy varying shown in the right of Fig.~\ref{fig:dis-analy}.

These results indicate that our {\modelshortname} encourages task-specific knowledge adaptation due to converging the ideal task-specific space. 
Besides, we observe two details. 
First, after epoch 1, CUS-CLIP's distance reduces by {\bf 2.2}, which is {\bf 58.6} time of TGT's decrease of {\bf 0.038}. 
This is because CLIP represents a heterogeneous space of vision-language, much different from the vision space $\mathcal{O}$. 
Furthermore, the large distance decrease confirms the effect of customization. 
Second, the synchronized distance reductions of CUS-CLIP and TGT indicate the interaction between the target model and CLIP is a crucial design for task-specific distillation.

\section{Conclusion}\label{sec:con}

We present an innovative approach, referred to as {\modelshortname}, designed to tackle the SFDA problem. To the best of our knowledge, this marks the initial endeavor to address SFDA by leveraging a pretrained ViL foundation model, departing from previous approaches that predominantly concentrated on self-mining auxiliary information.
{\modelshortname} is featured with alternating between customization of the ViL model and the transfer of task-specific knowledge from the customized ViL model. We introduce two pivotal designs: a mutual information-based alignment for ViL customization and a most-likely category encouragement for more precise adaptation of task-specific knowledge.
Our method's effectiveness is validated by state-of-the-art experimental results across four challenging datasets.

\section*{Acknowledgement}\label{sec:con}
This work is supported by the German Research Foundation (DFG) and National Natural Science Foundation of China (NSFC) in project Crossmodal Learning under contract Sonderforschungsbereich Transregio 169, the Hamburg Landesforschungsf{\"o}rderungsprojekt Cross, NSFC (61773083); NSFC (62206168, 62276048). Shanghai SAST Funding (202410013).

\clearpage
\setcounter{page}{1}
\maketitlesupplementary


\section{A Proof of Theorem~1.} 
\noindent\textbf{Restatement of Theorem~1} 
\textit{Given two random variables $X$, $Y$. Their mutual information ${\rm{I}}\left( X, Y \right)$ and KL divergence $D_{\rm{KL}}\left(X || Y\right)$ satisfy the unequal relationship as follows.} 
\begin{equation}
    \label{eqn:dsib}
    -{\rm{I}}\left( X, Y \right) \leq D_{\rm{KL}}\left( X || Y \right). 
\end{equation} 
\label{thm-one}
{\bf \textit{Proof.}}
Suppose the probability density function (PDF) of $X$ and $Y$ are $p(\boldsymbol{x})$ and $p(\boldsymbol{y})$, respectively; their join PDF is $p(\boldsymbol{x},\boldsymbol{y})$. 
We have  
\begin{equation*}
    \begin{split}
    {\rm{I}}\left(X, Y\right) 
    &= \sum p(\boldsymbol{x}, \boldsymbol{y})\log\frac{p(\boldsymbol{x},\boldsymbol{y})}{p(\boldsymbol{x})\cdot p(\boldsymbol{y})} \\
    &= D_{\rm{KL}}\left( p(\boldsymbol{x},\boldsymbol{y})~||~p(\boldsymbol{x})\cdot p(\boldsymbol{y}) \right). \\
    \end{split}
\end{equation*} 
Well known, the KL divergence is non-negative~\cite{eguchi2006interpreting}. 
Thus, 
\begin{equation*}
-{\rm{I}}\left( X, Y \right) \leq 0 \leq D_{\rm{KL}}\left( X || Y \right)
\end{equation*}

\section{Evaluation Datasets}
We evaluate four standard benchmarks below.
\begin{itemize}
    \item \textbf{Office-31}~\cite{saenko2010adapting} is a small-scaled dataset including three domains, i.e., Amazon~(A), Webcam~(W), and Dslr~(D), all of which are taken of real-world objects in various office environments. The dataset has 4,652 images of 31 categories in total. Images in (A) are online e-commerce pictures. (W) and (D) consist of low-resolution and high-resolution pictures. 

    \item \textbf{Office-Home}~\cite{venkateswara2017deep} is a medium-scale dataset that is mainly used for domain adaptation, all of which contains 15k images belonging to 65 categories from working or family environments. The dataset has four distinct domains, i.e., Artistic images~(Ar), Clip Art~(Cl), Product images~(Pr), and Real-word images~(Rw). 
    
    \item \textbf{VisDA}~\cite{peng2017visda} is a challenging large-scale dataset with 12 types of synthetic to real transfer recognition tasks. The source domain contains 152k synthetic images~(Sy), whilst the target domain has 55k real object images~(Re) from the famous Microsoft COCO dataset.

    \item \textbf{DomainNet-126}~\cite{peng2019moment} is another large-scale dataset. As a subset of DomainNet containing 600k images of 345 classes from 6 domains of different image styles, this dataset has 145k images from 126 classes, sampled from 4 domains, Clipart~(C), Painting~(P), Real~(R), Sketch~(S), as ~\cite{saito2019semi} identify severe noisy labels in the dataset. 
\end{itemize}

\section{Implementation Details}

\paragraph{Souce model pre-training.}

For all transfer tasks on the three datasets, we train the source model $\theta_s$ on the source domain in a supervised manner using the following objective of the classic cross-entropy loss with smooth label, like other methods~\cite{liang2020we,yang2021nrc,tang2022sclm}.
\begin{equation}
    \label{equ:opt_source}
    \begin{split}
        L_{\rm{s}}\left(\mathcal{X}_s, \mathcal{Y}_s; \theta_s\right)=-\frac{1}{n_s}\sum_{i=1}^{n_s} \sum_{c=1}^{C}\tilde{{l}}_{i,c}^s\log{p}_{i,c}^s, \nonumber
    \end{split}
\end{equation}
where $n_s$ is the number of the source data, ${p}_{i,c}^s$ is the $c$-th element of $\boldsymbol{p}_i^s=\theta_s({\boldsymbol{x}}_{i}^{s})$ that is the category probability vector of input instance ${\boldsymbol{x}}_{i}^{s}$ after $\theta_s$ mapping;
$\tilde{{l}}_{i,c}^s$ is the $c$-th element of the smooth label~\citep{muller2019does} $\tilde{\boldsymbol{l}}_i^s=(1-\sigma){\kern 2pt}\boldsymbol{l}_i^s + \sigma/C$, in which $\boldsymbol{l}_i^s$ is a one-hot encoding of hard label $y_i^s$ and $\sigma=0.1$. 
The source dataset is divided into the training set and testing set in a 0.9:0.1 ratio.
\begin{table*}[t]
    \caption{Full results (\%) of Closed-set SFDA on \textbf{VisDA}. 
    \textbf{SF} and \textbf{M} mean source-free and multimodal, respectively.}
    \label{tab:vc}
    \scriptsize
    \renewcommand\tabcolsep{4pt}
    \renewcommand\arraystretch{1.05}
    \centering
    \begin{tabular}{ l l | c c| c c c c c c c c c c c c |c}
        \toprule
        Method &Venue &\textbf{SF}  &\textbf{M} &plane &bcycl &bus &car &horse &knife &mcycl &person &plant &sktbrd &train &truck &Perclass \\
        \midrule
        Source                        &-  &-  &-  &60.7 &21.7 &50.8 &68.5 &71.8 &5.4  &86.4 &20.2 &67.1 &43.3 &83.3 &10.6  &49.2 \\
        \midrule
        DAPL-RN~\cite{ge2022domain}      &TNNLS23  &\xmark  &\cmark &97.8 &83.1 &88.8 &77.9 &97.4 &91.5 &94.2 &79.7 &88.6 &89.3 &92.5 &62.0 &86.9 \\
        PADCLIP-RN~\cite{lai2023padclip} &ICCV23   &\xmark  &\cmark &96.7 &88.8 &87.0 &82.8 &97.1 &93.0 &91.3 &83.0 &95.5 &91.8 &91.5 &63.0 &88.5 \\
        ADCLIP-RN~\cite{singha2023ad}    &ICCVW23   &\xmark  &\cmark &\textbf{\color{cmred}98.1} &83.6 &\textbf{\color{cmred}91.2} &76.6 &\textbf{\color{cmred}98.1} &93.4 &\textbf{\color{cmred}96.0} &81.4 &86.4 &91.5 &92.1 &64.2 &87.7 \\
        \midrule
        SHOT~\cite{liang2020we}     &ICML20   &\cmark  &\xmark &95.0 &87.4 &80.9 &57.6 &93.9 &94.1 &79.4 &80.4 &90.9 &89.8 &85.8 &{57.5} &82.7 \\
        NRC~\cite{yang2021nrc}      &NIPS21   &\cmark  &\xmark &96.8 &{91.3} &82.4 &62.4 &96.2 &95.9 &86.1 &\textbf{\color{cmred}90.7} &94.8 &94.1 &90.4 &59.7 &85.9 \\
        GKD~\cite{tang2021model}           &IROS21   &\cmark  &\xmark &95.3 &87.6 &81.7 &58.1 &93.9 &94.0 &80.0 &80.0 &91.2 &91.0 &86.9 &56.1 &83.0 \\
        AaD~\cite{yang2022attracting}      &NIPS22   &\cmark  &\xmark &97.4 &90.5 &80.8 &76.2 &97.3 &96.1 &89.8 &82.9 &95.5 &93.0 &92.0 &64.7 &88.0 \\
        AdaCon~\cite{chen2022contrastive}  &CVPR22   &\cmark  &\xmark &97.0 &84.7 &84.0 &77.3 &96.7 &93.8 &91.9 &84.8 &94.3 &93.1 &94.1 &49.7 &86.8 \\
        CoWA~\cite{lee2022confidence}      &ICML22   &\cmark  &\xmark &96.2 &89.7 &83.9 &73.8 &96.4 &\textbf{\color{cmred}97.4} &89.3 &86.8 &94.6 &92.1 &88.7 &53.8 &86.9 \\
        SCLM~\cite{tang2022sclm}       &NN22     &\cmark  &\xmark &97.1 &90.7 &85.6 &62.0 &97.3 &94.6 &81.8 &84.3 &93.6 &92.8 &88.0 &55.9 &85.3 \\
        ELR~\cite{yi2023source}        &ICLR23   &\cmark  &\xmark &97.1 &89.7 &82.7 &62.0 &96.2 &97.0 &87.6 &81.2 &93.7 &94.1 &90.2 &58.6 &85.8 \\
        PLUE~\cite{Litrico_2023_CVPR}  &CVPR23   &\cmark  &\xmark &94.4 &\textbf{\color{cmred}91.7} &89.0 &70.5 &96.6 &94.9 &92.2 &88.8 &92.9 &95.3 &91.4 &61.6 &88.3 \\
        TPDS~\cite{tang2023source}     &IJCV23   &\cmark  &\xmark &97.6 &91.5 &89.7 &83.4 &97.5 &{96.3} &92.2  &82.4 &\textbf{\color{cmred}96.0} &94.1 &90.9&{40.4} &87.6 \\
        \rowcolor{gray! 40} \textbf{\modelshortname}-C-RN   &- &\cmark &\cmark &97.7 &87.6 &90.5 &\textbf{\color{cmred}83.6} &96.7 &95.8 &94.8 &74.1 &92.4 &93.8 &92.9 &65.5 &88.8 \\
        \rowcolor{gray! 40} \textbf{\modelshortname}-C-B32  &- &\cmark &\cmark &97.5 &89.0 &90.8 &83.5 &97.8 &97.3 &93.2 &83.5 &95.2 &\textbf{\color{cmred}96.8} &\textbf{\color{cmred}93.7} &\textbf{\color{cmred}65.9} &\textbf{\color{cmred}90.3} \\
        \bottomrule
    \end{tabular}
\end{table*}

\begin{figure*}[t]
    \begin{center}
       \includegraphics[width=0.88\linewidth, height=0.28\linewidth]{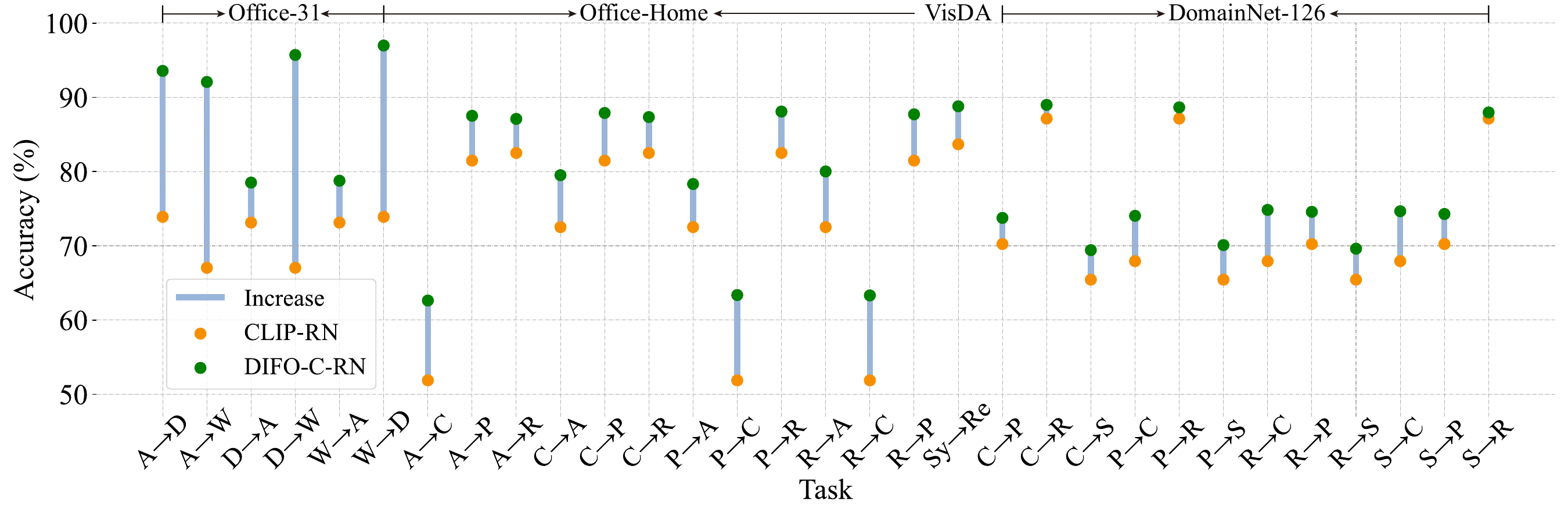}
       \includegraphics[width=0.88\linewidth, height=0.28\linewidth]{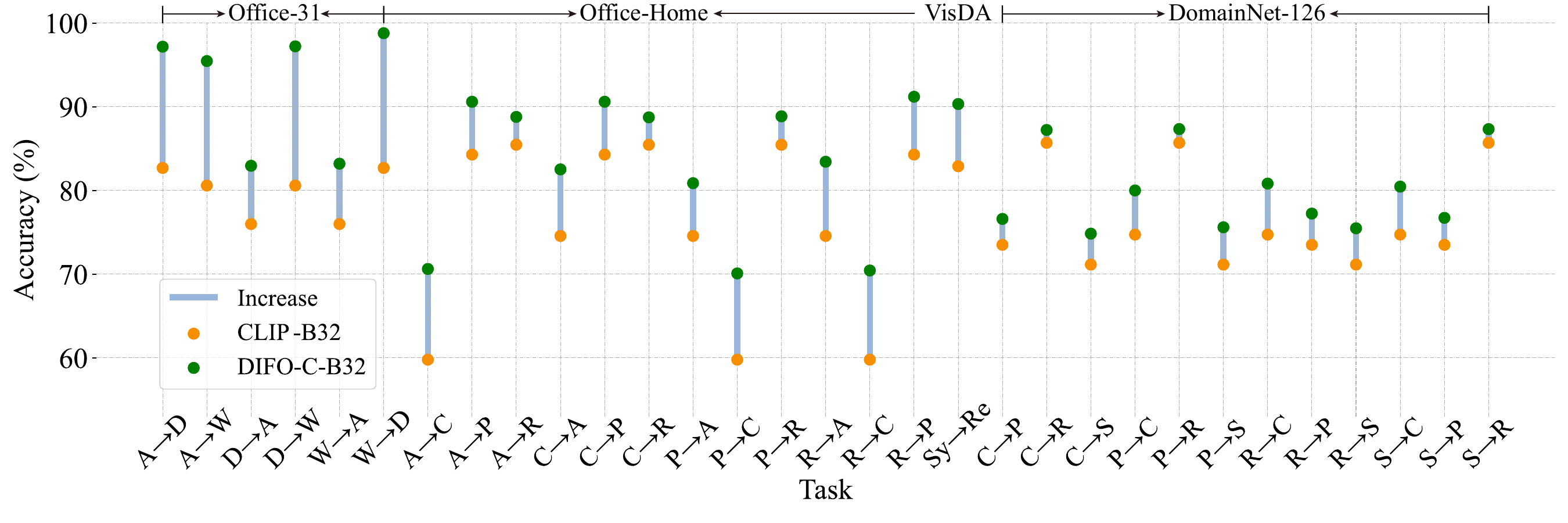}
    \end{center}
    \setlength{\abovecaptionskip}{0cm}
    \caption{Transfer performance comparison of {\bf \modelshortname} and CLIP on all tasks of the four evaluation datasets.
    {\bf Top:}  {\bf \modelshortname}-C-RN {\bf v.s.} CLIP-RN.
    {\bf Bottom:} {\bf \modelshortname}-C-B32 {\bf v.s.} CLIP-B32.} 
    \label{fig:coma-clip}
\end{figure*}

\begin{table*}[t]
    \caption{Full results~(\%) of Partial-set SFDA and Open-set SFDA on {\bf Office-Home}.}
    \label{tab:ob-ps-os}
    \renewcommand\tabcolsep{3.2pt} 
    \renewcommand\arraystretch{1.05} 
    \scriptsize
    \centering
    \begin{tabular}{ l l  c c c c c c c c c c c c c}
        \toprule
        Partial-set SFDA  &Venue  
        &Ar$\to$Cl &Ar$\to$Pr &Ar$\to$Rw
        &Cl$\to$Ar &Cl$\to$Pr &Cl$\to$Rw    
        &Pr$\to$Ar &Pr$\to$Cl &Pr$\to$Rw  
        &Rw$\to$Ar &Rw$\to$Cl &Rw$\to$Pr &Avg. \\
        \midrule
        Source          &--  &45.2 &70.4 &81.0 &56.2 &60.8 &66.2 &60.9 &40.1 &76.2 &70.8 &48.5 &77.3 &62.8 \\
        \midrule
        SHOT~\cite{liang2020we}        &ICML20   &64.8 &85.2 &\textbf{\color{cmred}92.7} &76.3 &77.6 &88.8 &79.7 &64.3 &89.5 &80.6 &66.4 &85.8 &79.3 \\
        HCL~\cite{huang2021model}      &NIPS21   &65.6 &85.2 &\textbf{\color{cmred}92.7} &77.3 &76.2 &87.2 &78.2 &66.0 &89.1 &81.5 &68.4 &87.3 &79.6 \\ 
        CoWA~\cite{lee2022confidence}  &ICML22   &69.6 &93.2 &92.3 &78.9 &81.3 &92.1 &79.8 &71.7 &90.0 &83.8 &\textbf{\color{cmred}72.2} &\textbf{\color{cmred}93.7} &83.2 \\
        AaD~\cite{yang2022attracting}  &NIPS22   &67.0 &83.5 &93.1 &80.5 &76.0 &87.6 &78.1 &65.6 &90.2 &83.5 &64.3 &87.3 &79.7 \\
        CRS~\cite{zhang2023class}      &CVPR23   &68.6 &85.1 &90.9 &80.1 &79.4 &86.3 &79.2 &66.1 &90.5 &82.2 &69.5 &89.3 &80.6 \\
        \rowcolor{gray! 40} \textbf{\modelshortname}-C-B32   &--   
        &\textbf{\color{cmred}70.2} &\textbf{\color{cmred}91.7} &91.5 
        &\textbf{\color{cmred}87.8} &\textbf{\color{cmred}92.6} &\textbf{\color{cmred}92.9} &\textbf{\color{cmred}87.3} &\textbf{\color{cmred}70.7} 
        &\textbf{\color{cmred}92.9} &\textbf{\color{cmred}88.5} &69.6 &91.5 &\textbf{\color{cmred}85.6} \\
        \hline
        \hline
        Open-set SFDA  &Venue  
        &Ar$\to$Cl &Ar$\to$Pr &Ar$\to$Rw
        &Cl$\to$Ar &Cl$\to$Pr &Cl$\to$Rw    
        &Pr$\to$Ar &Pr$\to$Cl &Pr$\to$Rw  
        &Rw$\to$Ar &Rw$\to$Cl &Rw$\to$Pr &Avg. \\
        \midrule
        Source          &-- &36.3 &54.8 &69.1 &33.8 &44.4 &49.2 &36.8 &29.2 &56.8 &51.4 &35.1 &62.3 &46.6 \\
        \midrule
        SHOT~\cite{liang2020we}       &ICML20   &64.5 &80.4 &84.7 &63.1 &75.4 &81.2 &65.3 &59.3 &83.3 &69.6 &64.6 &82.3 &72.8 \\
        HCL~\cite{huang2021model}     &NIPS21   &64.0 &78.6 &82.4 &64.5 &73.1 &80.1 &64.8 &59.8 &75.3 &\textbf{\color{cmred}78.1} &\textbf{\color{cmred}69.3} &81.5 &72.6 \\ 
        CoWA~\cite{lee2022confidence} &ICML22   &63.3 &79.2 &85.4 &67.6 &\textbf{\color{cmred}83.6} &82.0 &66.9 &56.9 &81.1 &68.5 &57.9 &\textbf{\color{cmred}85.9} &73.2 \\
        AaD~\cite{yang2022attracting} &NIPS22   &63.7 &77.3 &80.4 &66.0 &72.6 &77.6 &69.1 &\textbf{\color{cmred}62.5} &79.8 &71.8 &62.3 &78.6 &71.8 \\
        CRS~\cite{zhang2023class}     &CVPR23   &\textbf{\color{cmred}65.2} &76.6 &80.2 &66.2 &75.3 &77.8 &70.4 &61.8 &79.3 &71.1 &61.1 &78.3 &73.2 \\
        \rowcolor{gray! 40} \textbf{\modelshortname}-C-B32   &--  &64.5 &\textbf{\color{cmred}86.2} &\textbf{\color{cmred}87.9} &\textbf{\color{cmred}68.2} 
        &79.3 &\textbf{\color{cmred}86.1} &\textbf{\color{cmred}67.2} &62.1 &\textbf{\color{cmred}88.3} &71.9 &65.3 &84.4 &\textbf{\color{cmred}75.9} \\
        \bottomrule
    \end{tabular}
\end{table*}

\begin{figure*}[t]
        \begin{minipage}{0.48\linewidth}
	\begin{center}
		{\includegraphics[width=0.85\linewidth]{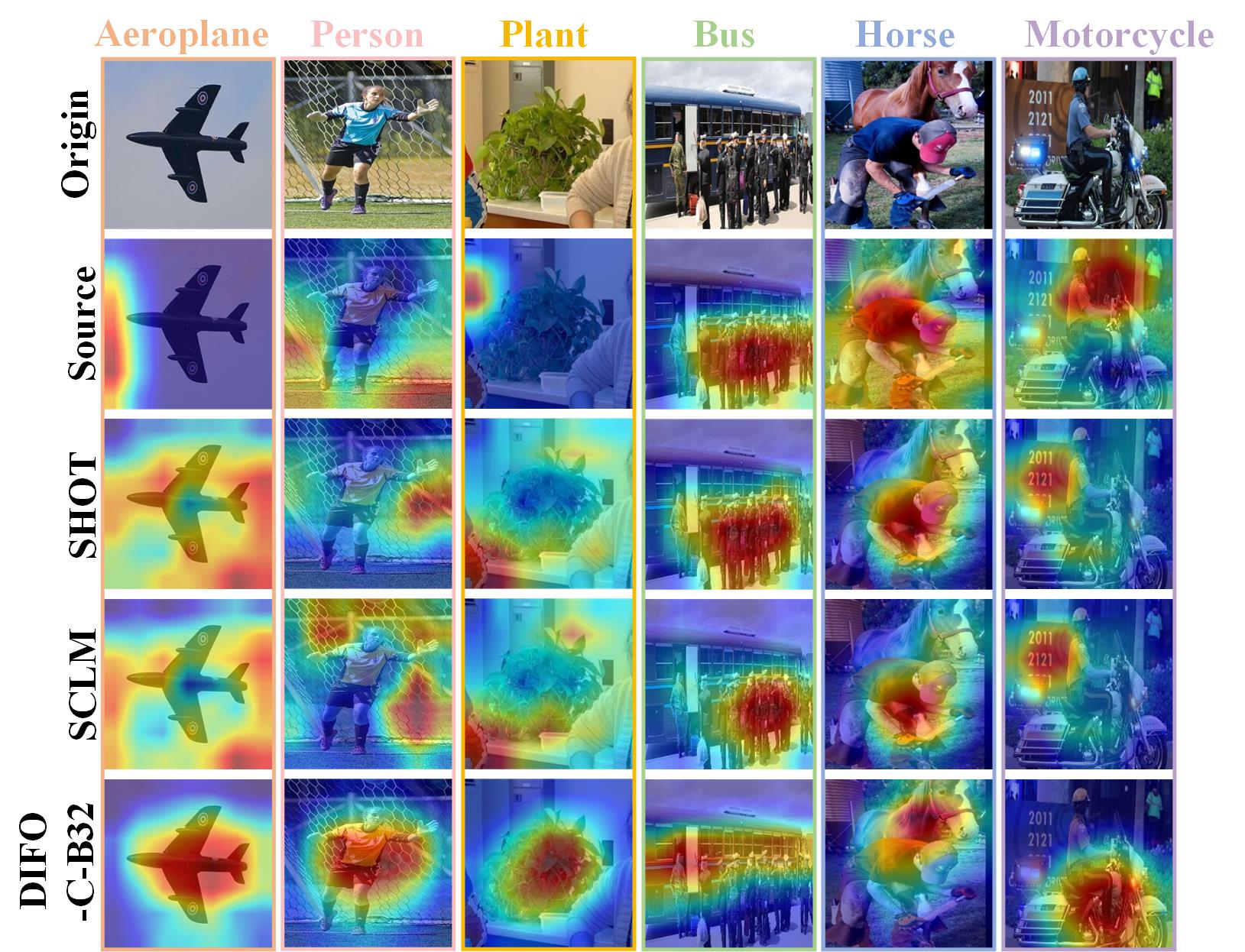}}
	\end{center}
	\caption{
        Grad-CAM visualization of {\bf \modelshortname}-C-B32 and typical comparison methods on toy samples selected from VisDA.}
	\label{fig:cam-comp}
        \end{minipage}
        \hfill
        \begin{minipage}{0.48\linewidth}
        \begin{center}
        {\includegraphics[width=0.85\linewidth]{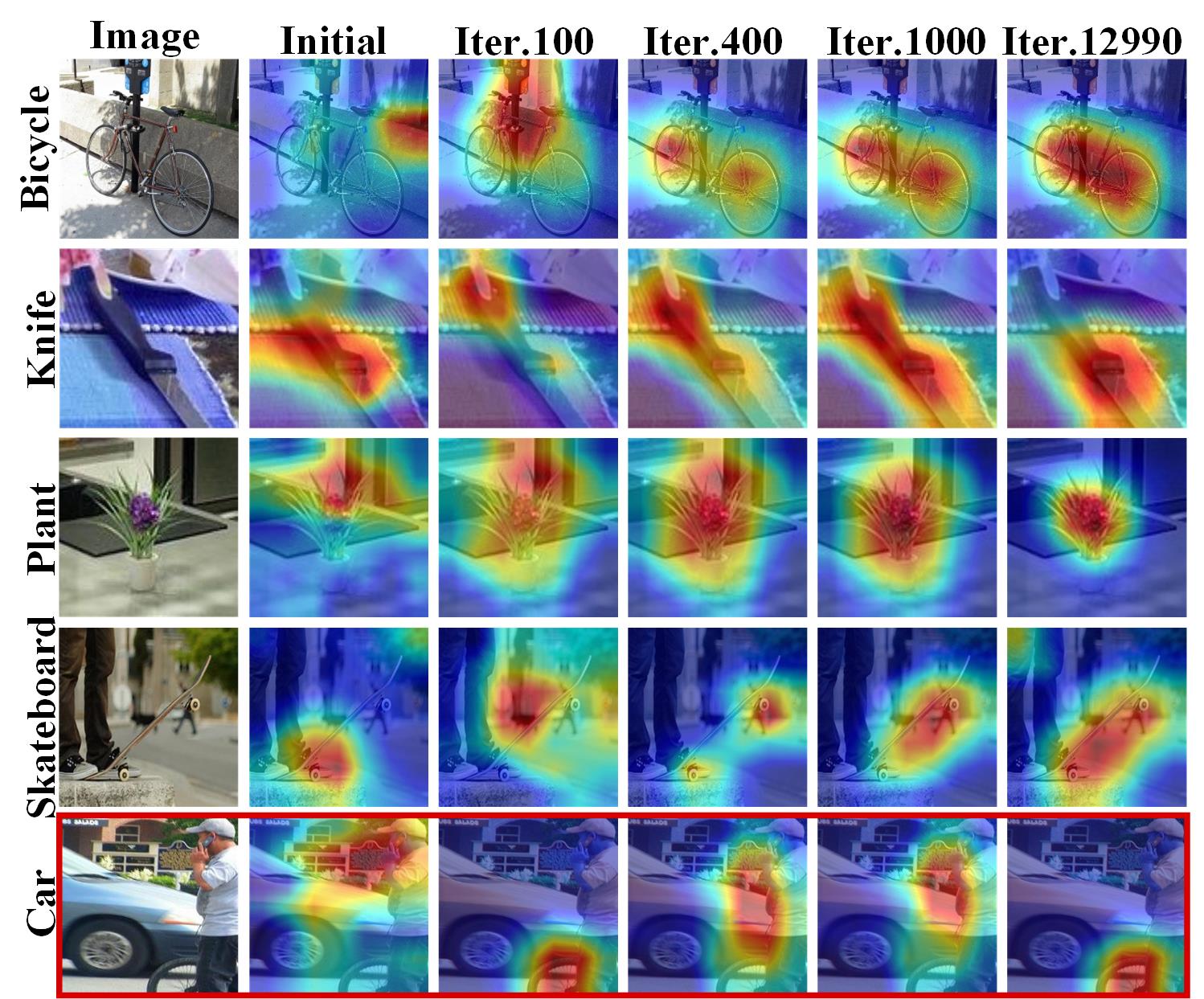}}
	\end{center}
        \caption{
        The evolving dynamics of model learning attention based on {\bf \modelshortname}-C-B32. The red bounding box indicates the failure case.}
	\label{fig:cam-dys}
        \end{minipage}
\end{figure*}

\vspace{5pt}
\noindent\textbf{Network setting.}
The {\modelshortname} model contains two network branches.
In the target model branch, the feature extractor consists of a deep architecture and a fully-connected layer followed by a batch-normalization layer. 
Same to the previous work~\cite{long2018conditional,xu2019larger,liang2020we,yang2021nrc,roy2022uncertainty}, the deep architecture is transferred from the deep models pre-trained on ImageNet (i.e., ResNet-50 is used on {\bf Office-31}, {\bf Office-Home} and {\bf DomainNet-126}, whilst ResNet-101 is adopted on {\bf VisDA}). 
The ending classifier is a fully-connected layer with weight normalization. 
On the other hand, the ViL model branch chooses the most adopted CLIP as the implementation where the text encoder's transformer-based architecture follows modification proposed in ~\cite{radford2021learning} as the backbone. 
Regarding the image encoder, we adopt two versions corresponding to the two implementations of {\modelshortname} in this paper, including  {\modelshortname}-C-B32 and   {\modelshortname}-C-RN. 
Specifically, in {\modelshortname}-C-B32, image encoders follow ViT-B/32 architecture proposed in CLIP~\cite{radford2021learning} while {\modelshortname}-C-RN uses  ResNet~\cite{he2016deep} as the backbone. 
The same as the target model mentioned above,  ResNet-101 is adopted on {\bf VisDA} and ResNet-50 is used on the rest datasets.


\vspace{5pt}
\noindent\textbf{Parameter setting.} 
For the trade-off parameter $\alpha$ and $\beta$ in the objective $L_{\rm{PC}}$ (Eq.~(6)) and $L_{\rm{MKA}}$ (Eq.~(7)) is set to 1.0 and 0.4 on all datasets, respectively.
The parameter of Exponential distribution $\lambda$ in Eq.~(4) is specified to 10.0. 
The temperature parameters in Eq.~(5) are $\tau=0.1$. 
The number of the most-likely categories is set to $N=2$. 


\vspace{5pt}
\noindent\textbf{Training setting.}
We adopt the batch size of 64, SGD optimizer with momentum 0.9 and 15 training epochs on all datasets. 
The prompt template for initiation is the mostly used \textit{'a photo of a [CLASS].'}~\cite{radford2021learning} where [CLASS] stands for the class name. 
All experiments are conducted with PyTorch on a single GPU of NVIDIA RTX.


\section{Supplementation of Full Experiment Results}  


\paragraph{Full results on VisDA.} 
As the supplement of results on VisDA, Tab.~\ref{tab:vc} presents the full classification details over the 12 categories. 
It is seen that {\modelshortname}-C-RN and {\modelshortname}-C-B32 obtain the best results in 7/12 categories compared with SFDA methods. 
Meanwhile, {\modelshortname}-C-RN and {\modelshortname}-C-B32 are on top of the second best UDA results in 8/12 categories.
Also, we note that the UDA method of ADCLIP beats {\modelshortname}-C-RN and {\modelshortname}-C-B32 on four transfer tasks. 
It is understandable that ADCLIP use the labelled source data, whilst our method cannot access the source data.
Despite this, {\modelshortname} still presents advantages over these source data-required method (see the average accuracy).

\vspace{5pt}
\noindent\textbf{Full results of comparison to CLIP.} 
As the supplementation of these domain-grouped results reported in the paper, Fig.~\ref{fig:coma-clip} gives a comprehensive visualization comparison with CLIP in the perspective of all 31 transfer tasks on the four evaluation datasets.  
It is seen that the results of {\modelshortname}~(marked by green circles) are above CLIP~(marked by orange circles) on all tasks, whether we use {\modelshortname}-C-RN or {\modelshortname}-C-B32. 

\vspace{5pt}
\noindent\textbf{Full results of Partial-set and Open-set SFDA.} 
As the supplementation of these average results in Tab.~5, Tab.~\ref{tab:ob-ps-os} gives the full classification accuracy over 12 transfer tasks in the {\bf Office-Home} dataset. 
As the top in Tab.~\ref{tab:ob-ps-os}, {\modelshortname}-C-B32 obtains best results on 9/12 tasks in the Partial-set SFDA and on the half tasks in the Open-set SFDA.


\section{Expanded Model Analysis} 


\paragraph{Grad-CAM visualization.} 
In Fig.~\ref{fig:cam-comp}, we present the Grad-CAM visualization~\cite{selvaraju2017grad} comparison with the source model and two typical SFDA methods, SHOT and SCLM, based on self-supervised learning without ViL model help. 
For the single object-contained images (see 1$\sim$3 column), {\modelshortname}-C-B32's attention focuses on the target object, whilst other methods cover the entire image. 
Regarding the multi-object-contained images (see 4$\sim$6 column), {\modelshortname}-C-B32's attention is more consistent with the target semantics given by the real labels than other methods focusing on the wrong object. 
These results explain the effectiveness of {\modelshortname}-C-B32 integrating the domain generality of the  ViL model and the task specificity of the source model.

\vspace{5pt}
\noindent\textbf{Attention-based evolving dynamics.}
To better understand the working of {\modelshortname}, this part visualizes the evolving dynamics of model learning attention during the training phase.
For a clear view, we display the Grad-CAM visualization results at some typical iterations, as shown in Fig.~\ref{fig:cam-dys}. 
Among the rightly classified images~(the top four rows), the attention smoothly concentrates to the discriminative visual patch. 
In contrast, the attention of the misclassified image~(the last row) converges to the meaningless one.

\begin{figure}[t]
    \begin{center}
        \subfigure[]{\includegraphics[width=0.48\linewidth,height=0.48\linewidth]{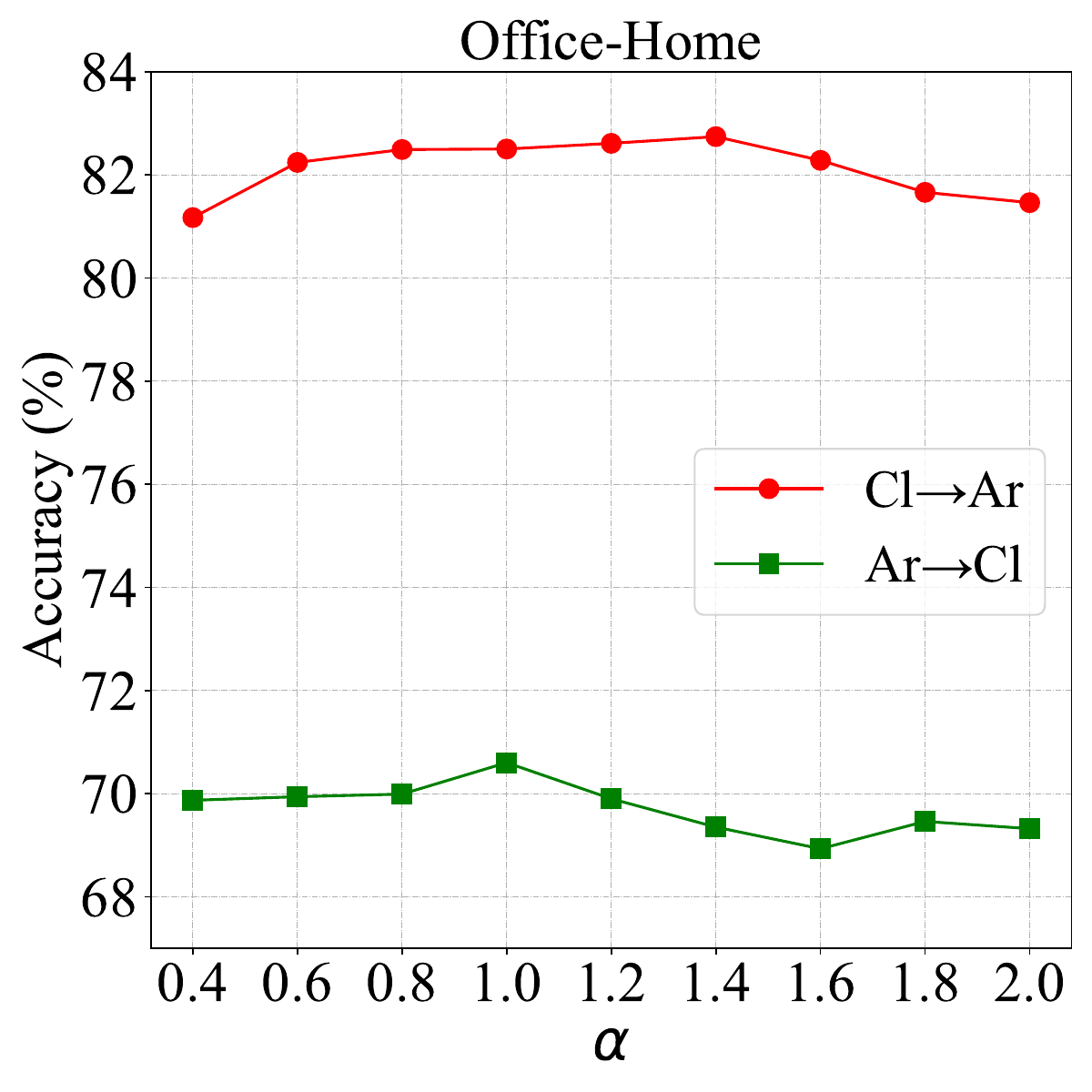}}~~
        \subfigure[]{\includegraphics[width=0.48\linewidth,height=0.48\linewidth]{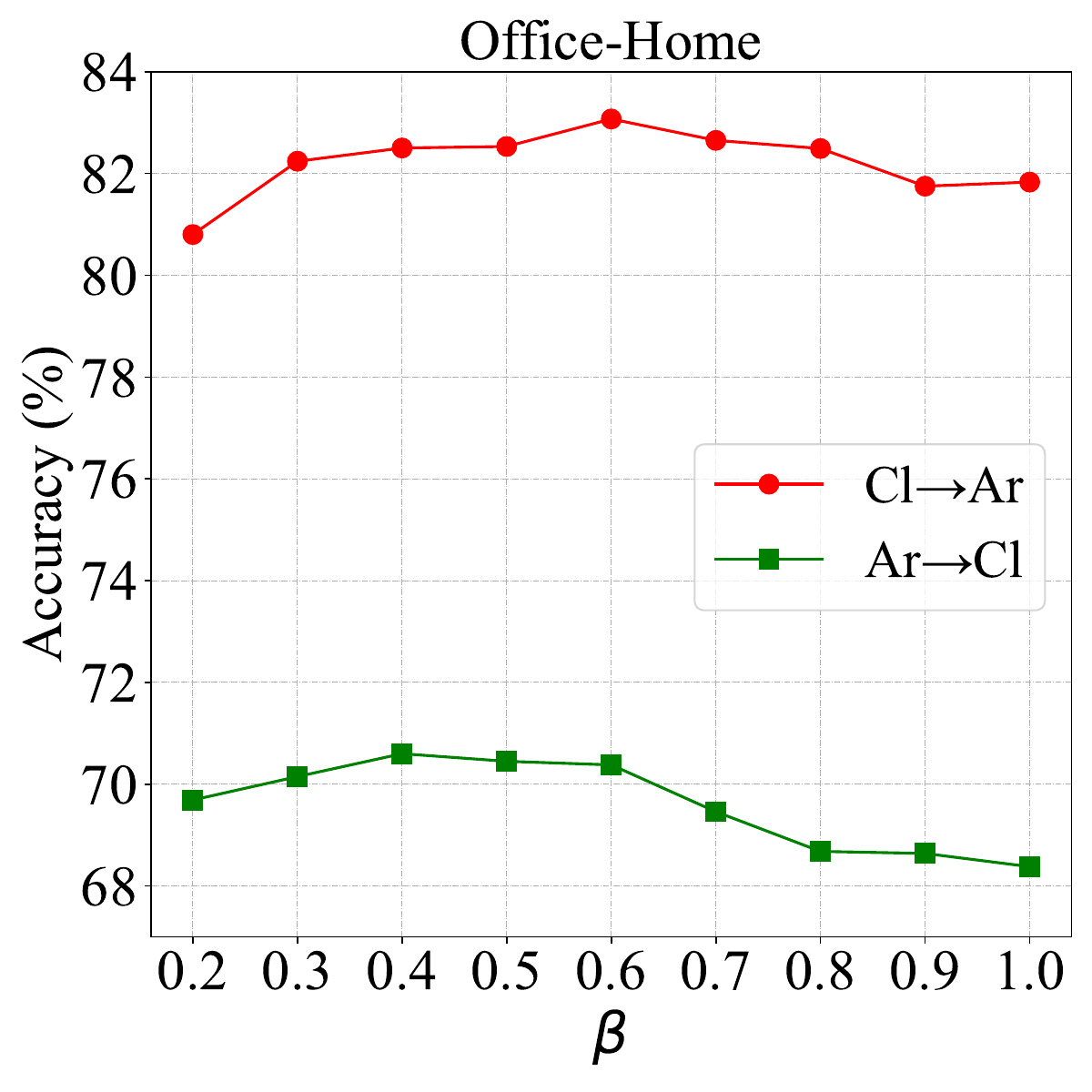}}~~\\
        \subfigure[]{\includegraphics[width=0.48\linewidth,height=0.48\linewidth]{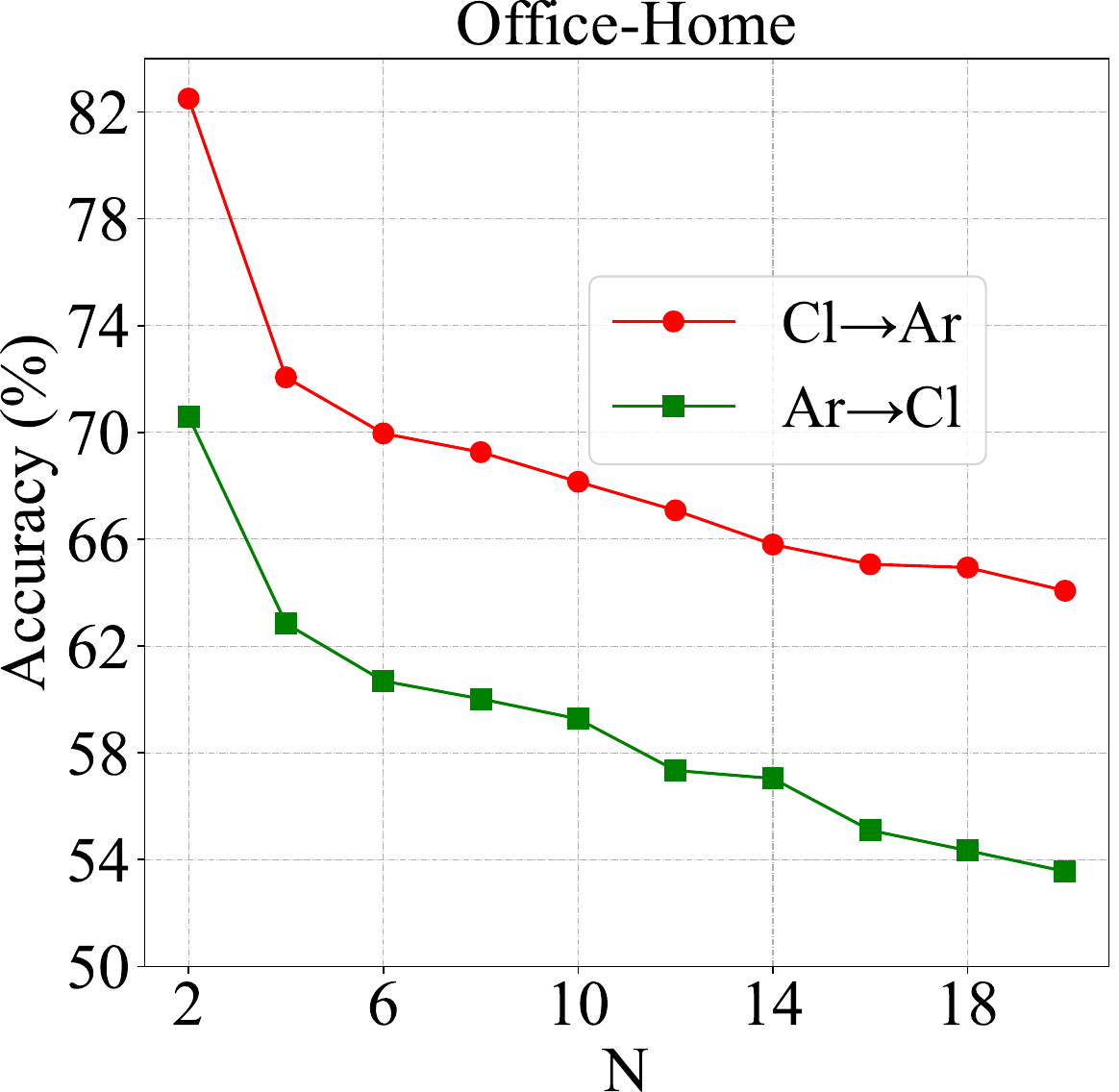}}~~
        \subfigure[]{\includegraphics[width=0.48\linewidth,height=0.48\linewidth]{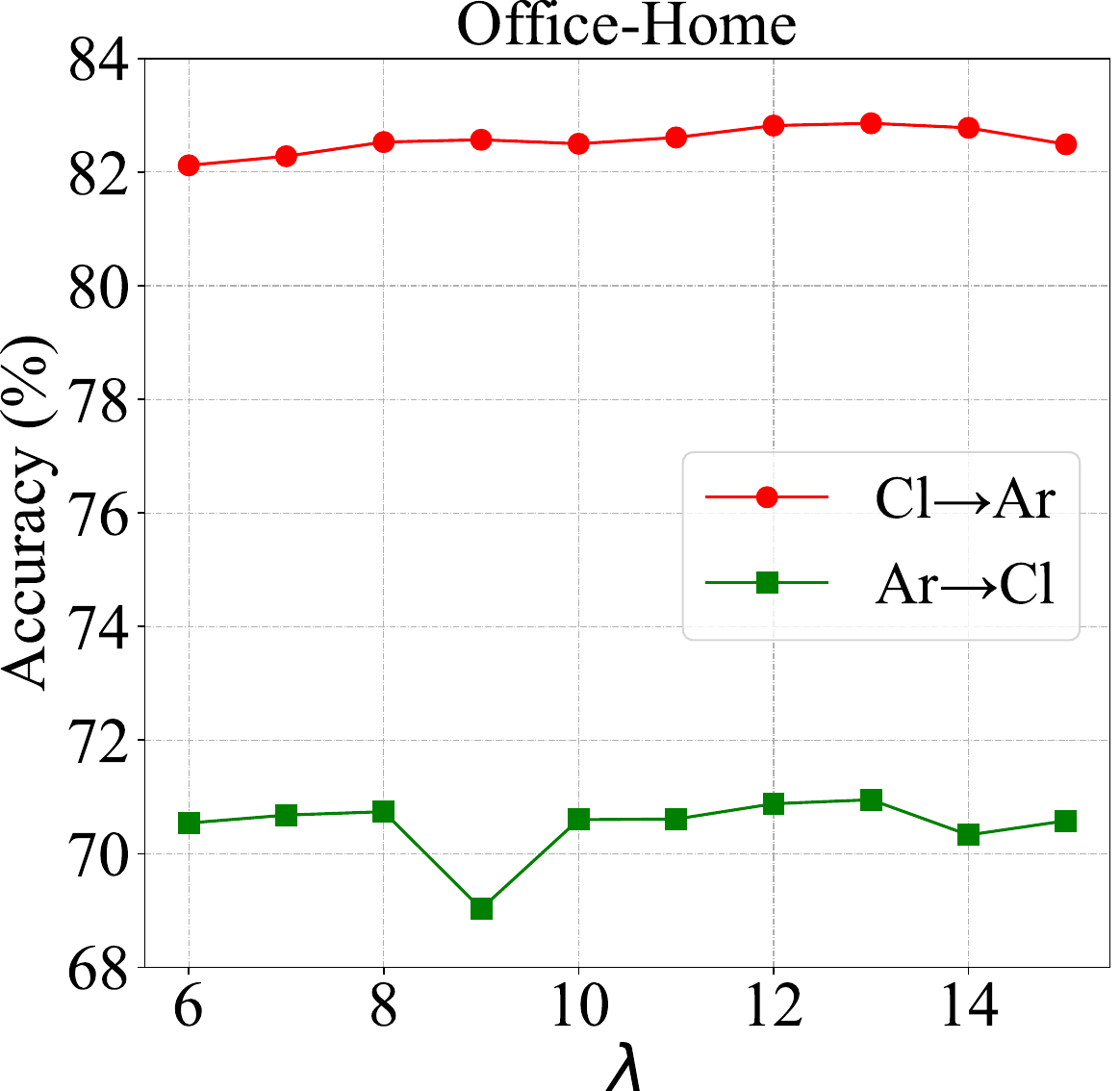}}~~
    \end{center}
    \setlength{\abovecaptionskip}{0cm}
    \caption{
    Performance sensitivity of the hyper-parameters. From (a) to (d), the four sub-figures present the accuracy changing as the parameters $\alpha$, $\beta$, $N$ and $\lambda$ varying, respectively.} 
    \label{fig:param}
\end{figure}
        
\begin{figure}[t]
    \begin{center}
    {\includegraphics[width=0.48\linewidth,height=0.48\linewidth]{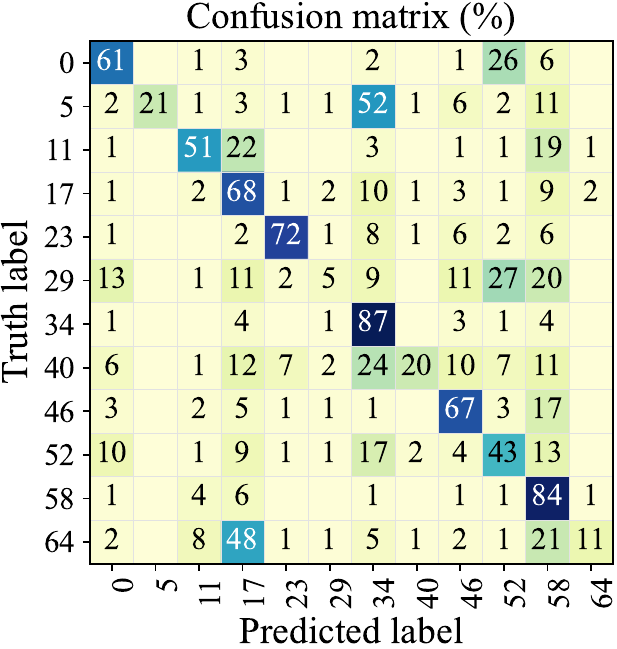}}
    {\includegraphics[width=0.48\linewidth,height=0.48\linewidth]{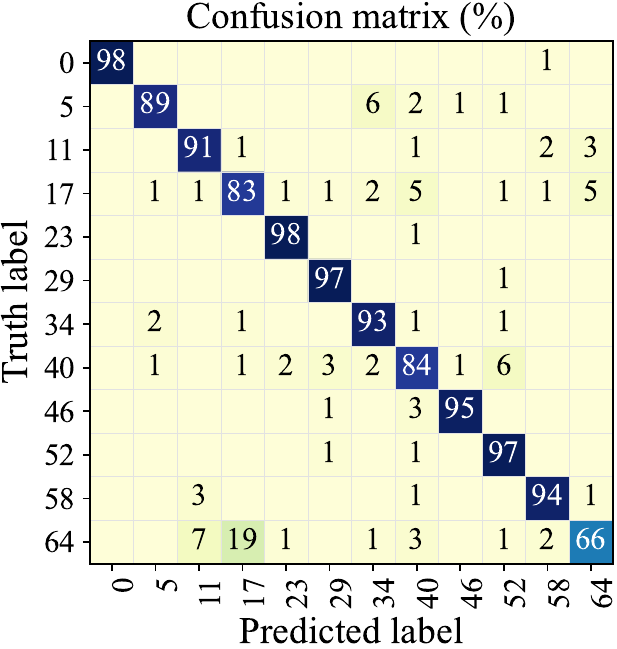}}
    \end{center}
    \caption{
    The confusion matrix for 12-way classification on {\bf VisDA}. 
    \textbf{Left:} Source model result, \textbf{Right:} {\modelshortname}-C-B32 result.}
\label{fig:fusionmtx}
\end{figure}

\begin{figure}[t]
    \begin{center}
        \includegraphics[width=0.7\linewidth]{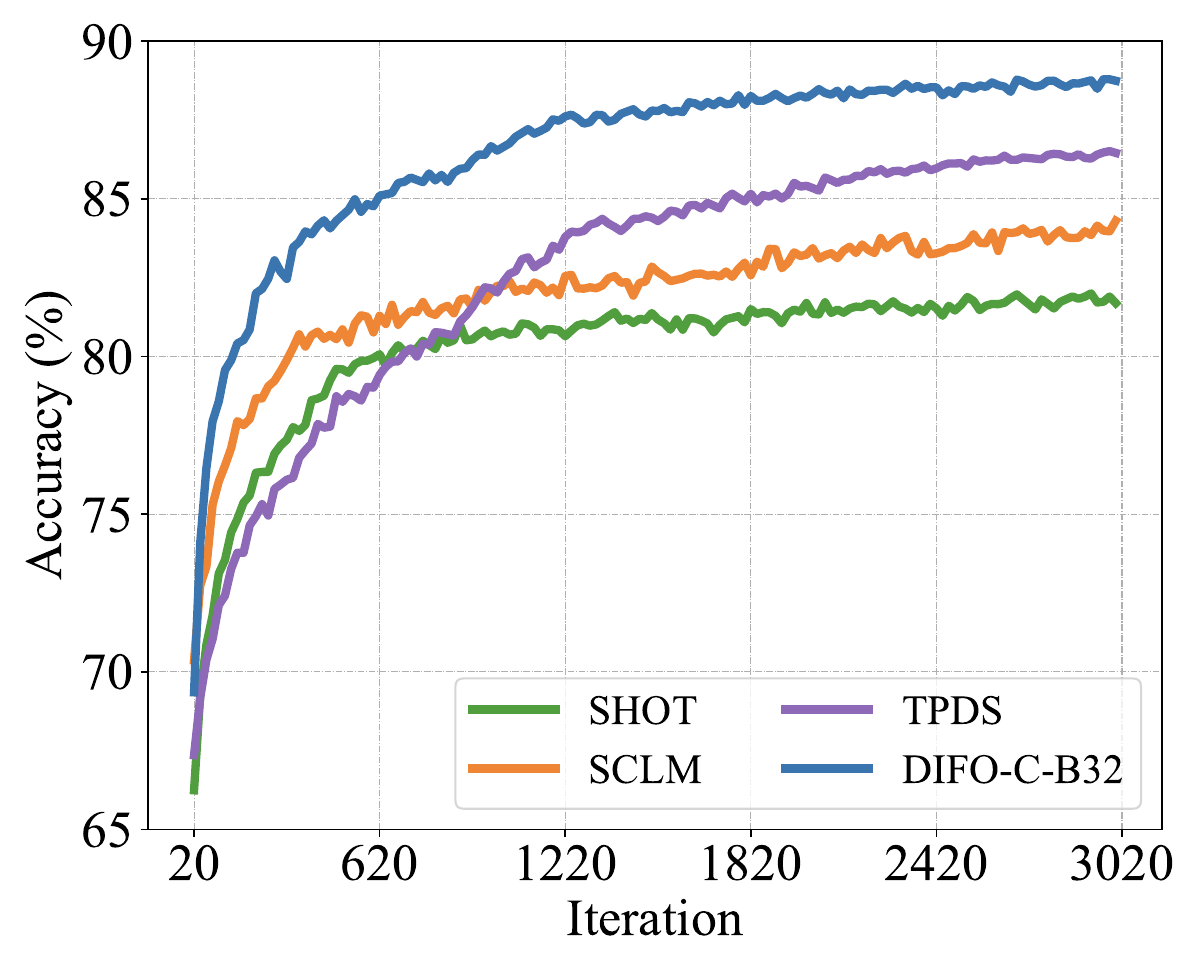}
    \end{center}
    \caption{Classification accuracy varying curve comparison on {\bf VisDA} during the adaptation phase.}
    \label{fig:stable}
\end{figure}

\vspace{5pt}
\noindent\textbf{Sensitivity of hyper-parameter.}
In the {\modelshortname} method, $\alpha$, $\beta$ are trade-off parameters in objective $L_{\rm{PC}}$~(see Eq.~(6)) and $L_{\rm{MKA}}$~(see Eq.~(7)).  
$\lambda$ is the parameter of Exponential distribution in Eq.~(4),  whilst $N$ is the number of the most-likely categories. 
This part discusses their performance sensitivity based on the symmetric transfer tasks Cl$\to$Ar and Ar$\to$Cl in the {\bf Office-Home} dataset. 
As depicted in Fig.~\ref{fig:param} (a), (b) and (d), when these parameters changes, there are no evident drops in the accuracy variation curves.  
This indicates that {\modelshortname} is insensitive to parameters $\alpha$, $\beta$ and $\lambda$. 
As for $N$, the accuracy gradually decreases as $N$ increases. 
This phenomenon is consistent with our expectation that small $N$ is better and a large value will introduce the semantic noise.

\vspace{5pt}
\noindent\textbf{Confusion matrix.}
To present a quantitative observation on the category, this part gives the confusion matrix based on the classification results on the {\bf VisDA} dataset.
For comparison, we show the confusion matrix of the source model at the left side of Fig.~\ref{fig:fusionmtx}. 
In the no-adaptation case, the misclassified data scatter over the matrix. 
After adaptation, the misclassified data are evidently corrected by {\modelshortname}-C-B32 at the right side of Fig.~\ref{fig:fusionmtx}.
It is seen that {\modelshortname}-C-B32 improves performance on all categories, and on some categories achieving significant growth. 
For instance, in the second category, the performance promotes by {\bf 68}\% (from {\bf 21}\% to {\bf 89}\%).

\vspace{5pt}
\noindent\textbf{Training stability.} 
Training stability is a vital characteristic of supervised learning methods. 
Based on the large-size dataset {\bf VisDA}, we present the adaptation details of {\modelshortname}-C-B32 using the accuracy varying curves on the target domain. 
For comparison, the curves of typical self-supervised methods, SHOT, SCLM and TPDS, are also depicted. 
As shown in Fig.~\ref{fig:stable}, the accuracy gradually increases to the maximum. 
This result confirms the training stability of {\modelshortname}-C-B32. 
Also, {\modelshortname}-C-B32 converges much faster than SHOT, SCLM and TPDS. 
It indicates that introducing task-specific knowledge from the ViL model is helpful in boosting the source model adaptation.

{
    \small
    \bibliographystyle{ieeenat_fullname}
    \bibliography{main}
}


\end{document}